\pdfoutput=1
\documentclass[a4paper,5p,onecolumn, fleqn]{cas-dc}

\usepackage[numbers]{natbib}

\usepackage{amsmath,amssymb}
\usepackage{graphicx}
\usepackage{textcomp}
\usepackage{multirow}
\usepackage{hyperref}
\usepackage{float}
\usepackage{xcolor,soul,framed} 
\sethlcolor{yellow} 
\usepackage{xcolor}

\usepackage{multicol}
\usepackage{subfig}
\usepackage{graphicx}
\usepackage{algorithmicx,algpseudocode}
\usepackage[linesnumbered, ruled, vlined]{algorithm2e}
\usepackage{orcidlink}

\usepackage{nomencl} 
\makenomenclature
\renewcommand*\nompreamble{\begin{multicols}{2}}
\renewcommand*\nompostamble{\end{multicols}}

\def\BibTeX{{\rm B\kern-.05em{\sc i\kern-.025em b}\kern-.08em
    T\kern-.1667em\lower.7ex\hbox{E}\kern-.125emX}}
    
\usepackage{threeparttable}
\usepackage{mathtools}

\newcolumntype{P}[1]{>{\centering\arraybackslash}p{#1}}

\def\tsc#1{\csdef{#1}{\textsc{\lowercase{#1}}\xspace}}
\tsc{WGM}
\tsc{QE}
\tsc{EP}
\tsc{PMS}
\tsc{BEC}
\tsc{DE}

\begin{document}


\shorttitle{Enhanced Encoder-Decoder Architecture for Accurate Monocular Depth Estimation}
\shortauthors{Das et~al.}

\title [mode=title]{Enhanced Encoder-Decoder Architecture for Accurate Monocular Depth Estimation}                      

\credit{Conceptualization of this study, Methodology, Software}

\address{Department of Computer Science and Engineering, Khulna University of Engineering {\&} Technology, Khulna - 9203, Bangladesh.}

\author{Dabbrata Das \orcidlink{0009-0008-0049-2048}}
\ead{dasdabbrata@gmail.com}

\author{Argho Deb Das \orcidlink{0009-0000-5756-2483}}
\ead{atdeb727@gmail.com}

\author{Farhan Sadaf \orcidlink{0009-0007-7550-7972}}
\cormark[1]
\ead{farhansadaf@cse.kuet.ac.bd}

\cortext[cor1]{Corresponding author}

\fntext[fn1]{Data: https://www.kaggle.com/datasets/soumikrakshit/nyu-depth-v2}
\fntext[fn2]{Code: https://github.com/dabbrata/Depth-Estimation-Enc-Dec}

\begin{abstract}
Estimating depth from a single 2D image is a challenging task due to the lack of stereo or multi-view data, which are typically required for depth perception. In state-of-the-art architectures, the main challenge is to efficiently capture complex objects and fine-grained details, which are often difficult to predict. This paper introduces a novel deep learning-based approach using an enhanced encoder-decoder architecture, where the Inception-ResNet-v2 model serves as the encoder. This is the first instance of utilizing Inception-ResNet-v2 as an encoder for monocular depth estimation, demonstrating improved performance over previous models. It incorporates multi-scale feature extraction to enhance depth prediction accuracy across various object sizes and distances. We propose a composite loss function comprising depth loss, gradient edge loss, and Structural Similarity Index Measure (SSIM) loss, with fine-tuned weights to optimize the weighted sum, ensuring a balance across different aspects of depth estimation. 
Experimental results on the KITTI dataset show that our model achieves a significantly faster inference time of 0.019 seconds, outperforming vision transformers in efficiency while maintaining good accuracy. On the NYU Depth V2 dataset, the model establishes state-of-the-art performance, with an Absolute Relative Error (ARE) of 0.064, a Root Mean Square Error (RMSE) of 0.228, and an accuracy of 89.3\% for $\delta$ < 1.25. These metrics demonstrate that our model can accurately and efficiently predict depth even in challenging scenarios, providing a practical solution for real-time applications.
\end{abstract}

\maketitle

\begin{table*}[!ht]   

\begin{framed}
\nomenclature{$DNN$}{Deep Neural Network}
\nomenclature{$RNN$}{Recurrent Neural Network}
\nomenclature{$LSTM$}{Long Short-Term Memory}
\nomenclature{$SGANVO$}{Stacked Generative Adversarial Network for Visual Odometry}
\nomenclature{$GANVO$}{Generative Adversarial Network for Visual Odometry}
\nomenclature{$GNN$}{Generative Adversarial Network}
\nomenclature{$CNN$}{Convolutional Neural Network}
\nomenclature{$RESNET$}{Residual Network}
\nomenclature{$IRv2$}{Inception Resnet V2}
\nomenclature{$IR-A$}{Inception Resnet A}
\nomenclature{$IR-B$}{Inception Resnet B}
\nomenclature{$IR-C$}{Inception Resnet C}
\nomenclature{$R-A$}{Reduction A}
\nomenclature{$R-B$}{Reduction B}
\nomenclature{$RELU$}{Rectified Linear Unit}
\nomenclature{$LR$}{Linear Regression}
\nomenclature{$ARE$}{Absolute Relative Error}
\nomenclature{$MARE$}{Mean Absolute Relative Error}
\nomenclature{$RMSE$}{Root Mean Square Error} 
\nomenclature{$MSE$}{Mean Square Error}
\nomenclature{$LOG10$}{Logarithmic Base 10}
\nomenclature{$R^2$}{Coefficient of Determination} 
\nomenclature{$SSIM$}{Structural Similarity Index Measure}
\nomenclature{$ViT$}{Vision Transformer}
\nomenclature{$IoU$}{Intersection over Union}
\nomenclature{$FLOPs$}{Floating Point Operations}

\printnomenclature
\end{framed}

\end{table*}

\section{Introduction}
 \label{sec:intro}
Scene depth estimation is a crucial task in computer vision that significantly enhances machine perception and comprehension of three-dimensional (3D) environments. Depth estimate is crucial for various applications, including autonomous driving, robotic navigation, virtual reality (VR), and augmented reality (AR). In these fields, precise depth information helps machines to efficiently interact with the actual world, ensuring safe navigation, accurate object detection, and realistic interaction with virtual environments.

Traditional depth estimation techniques, such as stereo vision and active depth sensing using LiDAR \cite{intro_LiDAR} or structured light, have been widely used to generate depth maps. These methods are effective but often require specialized hardware setups, including multiple cameras or expensive depth sensors, which significantly increase the cost and complexity of the system.
For example, in autonomous driving, LiDAR systems provide highly accurate depth maps, but their high cost and resource requirements limit their broad deployment. This has increased demand for more scalable and cost-effective solutions, particularly in scenarios where only a single camera is available.

Monocular depth estimation, which derives depth from a singular 2D image, has emerged as an appealing alternative due to its simplicity and minimal technology prerequisites. It removes the need for stereo vision systems or depth sensors, making it compatible with various devices, including mobile phones and drones. This challenge is tricky because, without stereo information, accurately estimating depth from a single image becomes more difficult. Problems like telling apart objects of different sizes or figuring out if one object is blocking another make monocular depth estimation especially difficult.

Early studies on monocular depth estimation predominantly utilized hand-crafted features and geometric indicators such as vanishing points, shadows, and defocus. These methods worked well in simple environments but struggled in complex real-world scenes. With machine learning, models using techniques like Scale-Invariant Feature Transform (SIFT) \cite{intro_SIFT} and Conditional Random Fields (CRF) \cite{intro_CRF} improved depth map predictions by learning from data. However, these models still had trouble working well in different types of scenes because they relied too much on hand-made features.

The recent growth of deep learning has significantly improved monocular depth estimation. Convolutional Neural Networks (CNNs), which can learn complex patterns directly from raw image data, are replacing traditional handcrafted features. Models proposed by Kim et al. \cite{CNN7} and Laina et al. \cite{CNN9} demonstrate that deep neural networks can accurately predict dense, high-resolution depth maps with significant accuracy, even in challenging environments. Moreover, the integration of Generative Adversarial Networks (GANs) \cite{intro_GAN} and attention mechanisms has advanced the field even further. Qiao et al. \cite{intro_ScienceDirect2} introduce an innovative multi-stage depth super-resolution network that utilizes explicit high-frequency data from a transformer and implicit signals from the frequency domain to improve depth map reconstruction. Besides, Zhang et al. \cite{fingerprint_fe_1} leveraged deep learning architectures for decomposing, extracting, and refining features from data or images to separate individual fingerprint features. Furthermore, Yadav et al. \cite{intro_ISA-GAN} presented an inception-based self-attentive generative adversarial network designed for high-quality facial image synthesis. The parallel self-attention module improves image quality by preserving spatial characteristics and speeds up convergence.

This paper focuses on advancing monocular depth estimation using deep learning, proposing a novel architecture based on Inception-ResNet-v2 \cite{networkarchitecture_encoder_IRv2}, which succeeds at capturing multi-scale features and refining depth predictions. Our contributions include a customized loss function which includes depth loss, gradient edge loss, and SSIM loss to optimize both the accuracy and structural consistency of depth maps. Here SSIM, edge, and depth losses are combined, functioning like feature fusion \cite{rgbt_tracking}, to create a robust model for depth prediction under varying conditions. We demonstrate the efficacy of our methodology through experiments on the NYU Depth V2 \cite{dataset_NYUv2}, KITTI \cite{dataset_KITTI} and Cityscapes \cite{dataset_cityscapes} datasets, where our model surpasses leading encoder-decoder techniques in accuracy on NYU Depth V2 and outperforms vision transformers in efficiency on KITTI.

The remaining parts of this work are structured as follows: Section II examines pertinent literature in monocular depth estimation, emphasizing significant progress in deep learning. Section III presents the approach and model's architecture, while Section IV covers the dataset, training procedure, and specifics of implementation. Section V includes the results and analysis, while Section VI finishes with future prospects and problems in the field.

\section{Related Works} \label{sec:related works}
Monocular depth estimation has made tremendous advances, transitioning from classic manual methods to advanced deep learning-based systems. This section examines major advances, focusing on early approaches, CNNs, and more recent innovations such as transformer-based systems, emphasizing their merits and limits.

\subsection{Traditional Methods: Early Foundations in Depth Estimation}
The first attempts at depth estimation were mostly based on classic computer vision techniques like stereo vision and structure-from-motion. These monocular depth estimation approaches used handmade features to estimate depth from two-dimensional images. Models such as the Scale-Invariant Feature Transform (SIFT) \cite{intro_SIFT} and Conditional Random Fields (CRF) \cite{intro_CRF} used predetermined features to estimate depth.

While these algorithms performed well in controlled environments, they were unable to effectively generalize to complex, real-world scenes due to their heavy dependence on static features and inability to adjust to changes in lighting, texture, or object shadowing.  For example, regular stereo vision systems require specialized hardware setups using many cameras or depth sensors such as LiDAR, making them expensive and difficult to implement on a large scale. Furthermore, these algorithms usually fail to adapt in dynamic or crowded situations, limiting their use in real-world applications like autonomous driving or augmented reality.

\subsection{The Deep Learning Era: Transition to data-driven approaches}
With the rise of deep learning, the field of monocular depth estimation made great progress. Convolutional Neural Networks (CNNs) had become popular as powerful tools to learn complex patterns from image data. 
Laina et al. \cite{CNN9} created a fully convolutional residual network for monocular depth estimation, incorporating an innovative upsampling method to enhance output resolution. Their model attained an Average Relative Error (ARE) of 0.127 on the NYU Depth V2 dataset \cite{dataset_NYUv2} and 0.176 on the Make3D dataset \cite{dataset_Make3D} respectively. 
Xu et al. \cite{CNN6} formulated a structured attention-guided conditional neural field model for the estimation of monocular depth. The integration of multi-scale characteristics and attention mechanisms resulted in superior performance compared to previous CRF-based models, achieving an ARE of 0.125 on the NYU Depth V2 dataset \cite{dataset_NYUv2} and 0.122 on the KITTI dataset \cite{dataset_KITTI}.

Zoran et al. \cite{CNN1} established a deep learning framework for mid-level vision tasks that acquires ordinal correlations among picture points. Utilizing the NYU Depth v2 \cite{dataset_NYUv2} dataset, their model achieved an RMSE(log) of 0.42 in depth estimation. 
Kim et al. \cite{CNN7} introduced a deep variational model for monocular depth estimation, integrating global and local predictions from two CNNs.Their model achieved an RMSE(log) of 0.172 on the NYU v2 \cite{dataset_NYUv2} dataset. 
Lee and Kim \cite{CNN5} created a convolutional neural network using an encoder-decoder architecture for depth estimation utilizing the NYUv2 \cite{dataset_NYUv2} dataset. Their approach achieved an RMSE(log) of 0.180, improving accuracy by effectively combining depth maps at multiple scales. 
Zhang et al. \cite{RW_dpsnet_geometry_reasoning} developed a multitask learning framework that estimates depth, camera posture, and semantic segmentation from monocular videos. Using geometric reasoning, their method achieved state-of-the-art performance with an RMSE of 6.317 on the KITTI dataset \cite{dataset_KITTI} and an average IoU improvement of 3.1\% on SYNTHIA \cite{dataset_synthia_dataset}. This highlights the effectiveness of combining different tasks into a single framework for increasing overall depth estimate performance.

Generative Adversarial Networks (GANs) also gained popularity during this period. Jung et al. \cite{GAN1} introduced Generative Adversarial Networks (GANs) for monocular depth estimation, employing a GlobalNet to extract global features and a RefinementNet to ascertain local structures from a single image, utilizing the NYU v2 \cite{dataset_NYUv2} dataset. Their method achieved significant improvement, with an ARE of 0.134. 
Similarly, Lore et al. \cite{GAN2} introduced a depth map estimation technique utilizing Conditional Generative Adversarial Networks (cGANs). The methodology was assessed using the NYU Depth v2 dataset \cite{dataset_NYUv2}, resulting in a root mean square error (RMSE) of 0.875. Their model outperformed traditional non-parametric sampling methods. 
Feng and Gu \cite{GAN3} devised an unsupervised methodology for depth and ego-motion estimation with Stacked GANs (SGANVO). The model surpassed current techniques in depth estimation on the KITTI dataset \cite{dataset_KITTI}, with an average RMSE log of 0.1623 across a number of scenarios. 

Recurrent Neural Networks (RNNs) were also studied for their ability to represent temporal dependencies in footage from videos. Kumar et al. \cite{RNN1} introduced an innovative convolutional LSTM-based \cite{RNN2} recurrent neural network architecture for monocular depth estimation from video sequences. Their methodology intended to take advantage of temporal dependencies between video frames, and they evaluate it using the KITTI dataset \cite {dataset_KITTI}. The best-performing model achieved an absolute relative error of 0.137. 
Mancini et al. \cite{RNN3} improved scene depth prediction by adding LSTM \cite{RNN2} units after the encoder network's convolutional layers. Their approach, as evaluated on the KITTI dataset \cite{dataset_KITTI}, significantly improved generalization, achieving an RMSE(log) of 0.366 and an Absolute Relative Difference of 0.312. 

Despite these advances, significant challenges were still present. Many deep learning algorithms require large labeled datasets for supervised training, which can be time-consuming and expensive to collect. Furthermore, the generalization of models across different and unknown environments remained a major challenge, especially for applications needing stability in highly dynamic environments such as outdoor and underwater environments. Another obstacle was computational resource needs, which limited the use of such models in resource-constrained systems such as mobile and embedded devices.

\begin{table*}[!ht, align=\flushleft, width=\textwidth]
    \caption{Summary of Impactful Works in Depth Estimation of Deep Learning}
    \centering
    \renewcommand{\arraystretch}{1.25}
    \footnotesize	
    \begin{threeparttable}
    \begin{tabular}{ P{2.2cm} P{2.0cm}  P{2.5cm} P{3.0cm} P{2.5cm} P{2cm} }
         \hline \hline
         \textbf{Reference} & \textbf{Analysis Type} & \textbf{Model} & \textbf{Technique} & \textbf{Dataset} & \textbf{Performance} \\
      \hline
     \noalign{\vskip 1mm}  
        Kumar et al. \cite{RNN1} & RNN-based & Convolutional LSTM & Utilized temporal dependencies between video frames & KITTI & Absolute Relative Error: 0.137 \\ 
      \noalign{\vskip 1mm} 
      \hline
      \hline
     
    \noalign{\vskip 1mm}  
        Jung et al. \cite{GAN1} & GAN-based & GlobalNet + RefinementNet & Extracted global features and estimated local structures & NYU Depth v2 & ARE: 0.134 \\
          \noalign{\vskip 1mm} 
      \hline 
    \noalign{\vskip 1mm}  
        Lore et al. \cite{GAN2} & GAN-based & Conditional GANs & Improved depth map estimation with cGANs & NYU Depth v2 & RMSE: 0.875 \\
   \noalign{\vskip 1mm}
      \hline    
      \hline
    \noalign{\vskip 1mm}  
        Li et al. \cite{CNN2} & CNN-based & VGG-16-based System & Fused depth and depth gradients & NYU Depth v2 & RMSE: 0.611 \\
 
   \noalign{\vskip 1mm} 
      \hline
      
   \noalign{\vskip 1mm}  
     Lee and Kim \cite{CNN5} & CNN-based & Encoder-Decoder CNN & Combined depth maps at multiple scales & NYUv2 and KITTI & RMSE(log): 0.180 on NYUv2 \\ 
      \noalign{\vskip 1mm} 
   \hline  
   
   \noalign{\vskip 1mm}  
     Xu et al. \cite{CNN6} & CNN-based & Structured Attention Model & Integrated multi-scale characteristics and attention mechanisms & NYU Depth v2 and KITTI & ARE: 0.125 on NYUv2 \\
      \noalign{\vskip 1mm} 
   \hline 
    \hline
    
   \noalign{\vskip 1mm}  
     Li et al. \cite{RW_binsformer_adaptive_bins} & Transformer-based & Adaptive Depth Prediction & Built depth bins and enhanced predictions across many scales & NYU, KITTI, and SUN RGB-D & RMSE(log): Improved up to 6.1\% on KITTI \\ 
      \noalign{\vskip 1mm} 
   \hline  

   \noalign{\vskip 1mm}  
     Lu and Chen \cite{RW_joint_self_supervised_depth_flow} & Hybrid & Self-supervised Depth + Optical Flow & Combined motion segmentation and self-supervised learning techniques & KITTI & RMSE(log): 0.1680 \\ 
      \noalign{\vskip 1mm} 
   \hline  \hline 
   
      \end{tabular}
      \end{threeparttable}
    \label{tab:depth_works}
\end{table*}

\subsection{Recent Innovations: Tackling Modern Challenges}
Recent innovations have addressed many of the shortcomings of previous techniques by adding unique architectures, self-supervised learning concepts, and task-specific optimizations. These developments have enhanced the accuracy and scalability of monocular depth estimating systems.

For instance, Li et al. \cite{RW_binsformer_adaptive_bins} proposed a new approach for determining depth from a single image using classification and regression algorithms. It use a Transformer-based technique to adaptively build depth bins and enhance predictions across many scales. It achieved state-of-the-art results, with RMSE(log) improvements of up to 6.1\% on KITTI \cite{dataset_KITTI} and significant generalization on different datasets.

Another recent contribution by Lu and Chen \cite{RW_joint_self_supervised_depth_flow} designed a framework for evaluating both depth and optical flow in dynamic conditions. Their method improves depth estimation accuracy by combining motion segmentation and self-supervised learning techniques. Tested on the KITTI dataset \cite{dataset_KITTI}, it obtained remarkable accuracy with metrics such as RMS(log) of 0.1680, outperforming previous approaches.

In the domain of aquatic environments, Lu and Chen \cite{RW_self_supervised_water_depth} proposed a self-supervised approach for estimating monocular depth in water scenarios with specular reflection priors. The model divides water surfaces and uses reflections as intra-frame supervision to calculate depth. When tested on the WRS dataset, it produced state-of-the-art results, including an Absolute Relative Error (AbsRel) of 0.121 and RMS of 4.43, surpassing previous approaches by up to 28\%.

Zhang et al. \cite{RW_pyrf_pcr_3d_pointcloud} proposed a three-step method for registering 3D point clouds in outside environments. It combines preprocessing, yaw angle estimation, and coarse registration with frequency histograms. In several outdoor circumstances, the approach outperformed state-of-the-art methods, reducing average angle errors by 62.8\% and translation errors by 46.5\%.

Zhou et al. \cite{RW_u_shaped_transformer_iqa} introduced a Perception-Oriented U-Shaped Transformer Network (U-Former) for assessing 360-degree image quality without using a reference image. The model extracts perceptual features using cube map projection, saliency-based self-attention, and a U-Former encoder. Zhou et al. \cite{RW_blind_iqa_self_attention} proposed a blind image quality assessment model that combines self-attention and recurrent neural networks. It collects local and global image features using windowed self-attention and GRUs, achieving cutting-edge performance on benchmark datasets while resolving drawbacks in prior CNN- and transformer-based BIQA approaches. These approaches offer ways to assess image quality while utilizing advanced transformer-based methods.

Furthermore, Xi et al. \cite{CNN11} proposed LapUNet, a deep-learning framework for monocular depth estimation. Their model introduced a Dynamic Laplacian Residual U-shape (DLRU) module and incorporated an ASPP module to enhance multi-scale contextual feature extraction. Using the NYU Depth V2 \cite{dataset_NYUv2} and KITTI datasets \cite{dataset_KITTI}, the model achieved a significant improvement in depth accuracy, with an RMSE of 0.406 on NYU Depth V2 \cite{dataset_NYUv2} and 2.247 on KITTI \cite{dataset_KITTI}. Song et al. \cite{CNN12} integrated CNN and Vision Transformer components, utilizing an improved HRFormer as the encoder. Using the KITTI \cite{dataset_KITTI}, Cityscapes, and Make3D datasets, MFFENet outperformed state-of-the-art methods, achieving an RMSE of 4.356 on KITTI \cite{dataset_KITTI}. Choudhary et al. \cite{CNN14} proposed a dual-channel convolutional neural network (MEStereo-Du2CNN) designed for robust depth estimation using multi-exposure stereo images. The model introduces a mono-to-stereo transfer learning approach and eliminates the traditional cost volume construction in stereo matching. On the Middlebury dataset \cite{dataset_Middlebury}, the model achieved an RMSE of 0.079.

These newest developments deal with key issues such as increasing depth prediction in complicated situations, maintaining fine-grained spatial information, and enabling improved generalization across various datasets. However, restrictions such as computing needs for real-time applications and dealing with extreme edge cases remain, providing room for further study. Table \ref{tab:depth_works} provides a summary of influential works in monocular depth estimation.\\

Building upon insights from previous studies and their associated challenges, it is evident that enhancing the encoder-decoder architecture can address many of these issues effectively. Our proposed model is designed to tackle these challenges. The key contributions of our work are summarized as follows:

\begin{itemize}
    \item We utilized the Inception-ResNet v2 (IRv2) architecture as the encoder, leveraging its simultaneous multi-scale feature extraction capability. This proved advantageous in achieving better accuracy, particularly for smaller and more complex objects.

    \item A composite loss function was employed, incorporating three distinct weights. After extensive experimentation with various combinations, the optimal values were identified, contributing to improved model performance.

    \item When compared to Vision Transformers (ViT), our IRv2-based model demonstrated significantly lower inference times. This efficiency enhancement makes it more suitable for real-time applications, offering a considerable advantage over ViT in practical scenarios.

    \item We experimented with our model in different indoor and outdoor scenarios, providing a detailed statistical analysis of the results, which revealed that our model outperforms several state-of-the-art architectures across various metrics.
\end{itemize}

\section{Methodology}
\label{sec:methodology}

This section provides a comprehensive explanation of data preprocessing and model architecture with detailed workflow and loss functions. 

\subsection{Preprocessing}
Preprocessing the images is a crucial step in building the model, as it ensures better results before passing the data into it.
We first applied the min-max normalization equation \eqref{eq:minmax}, a linear transformation that scales the image intensity values between 0 and 1. 
\begin{equation}
\bar{x} = \frac{x - x_{\min}}{x_{\max} - x_{\min}}
\label{eq:minmax}
\end{equation}
Here, the normalized value is ${\bar{x}}$, and the original value is x. Data augmentation, including horizontal flipping of images, is also applied during the preprocessing steps.

\begin{figure*}[h, align=\flushleft, width=\textwidth]
\centering
\includegraphics[width=\textwidth]{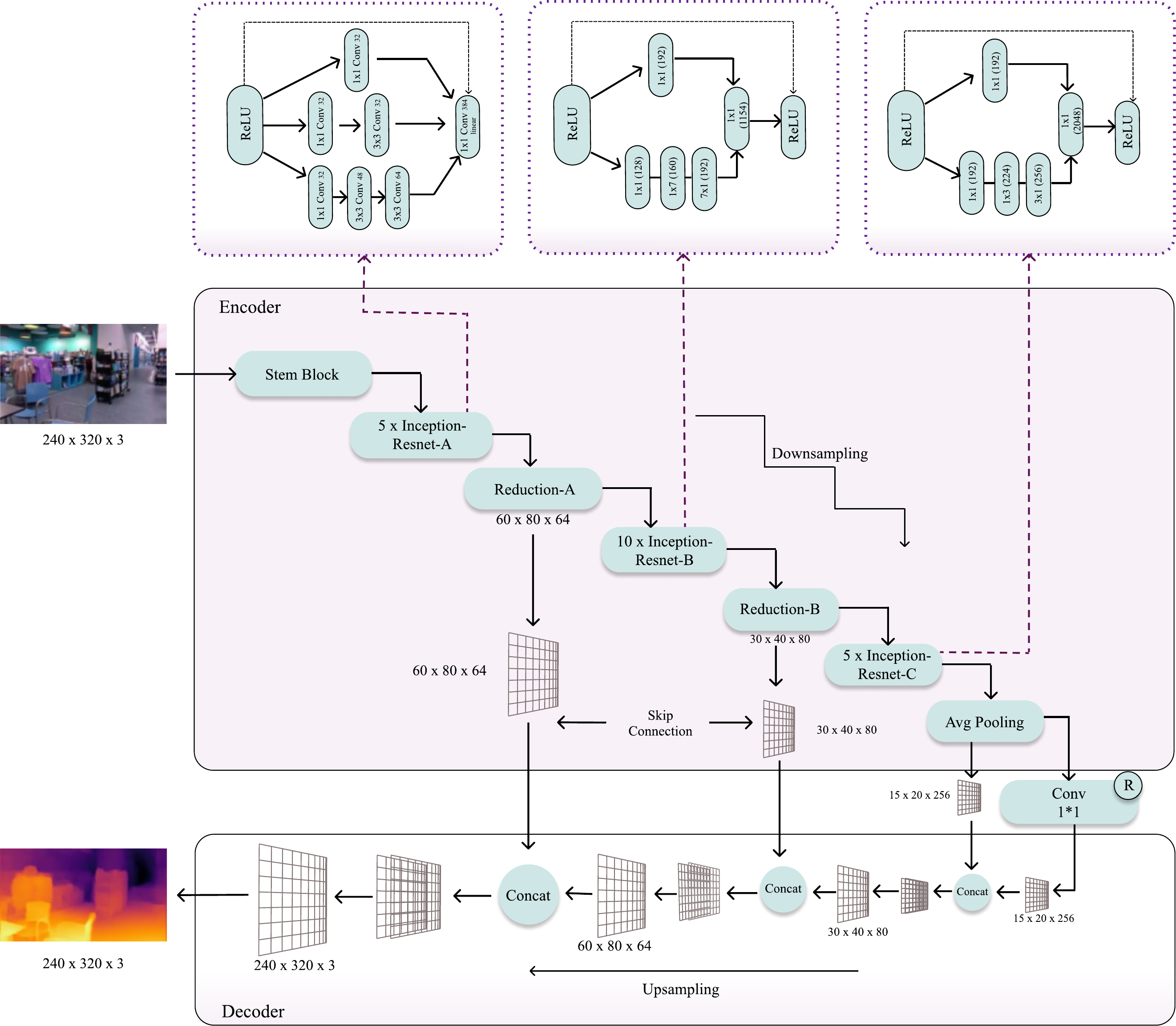}
\caption{\textbf{An outline of our network (encoder-decoder) architecture.} The encoder uses a pre-trained Inception-ResNet-v2 (IRv2) \cite{networkarchitecture_encoder_IRv2} network, consisting of several Inception-ResNet blocks (A, B, and C) and reduction layers. The decoder consists of convolutional layers that process the upsampled output from the previous layer, combined with the corresponding feature maps from the encoder.}
\label{fig1}
\end{figure*}

\subsection{Network Architecture}
Our model makes use of an encoder-decoder architecture based on deep convolutional neural networks, implemented using the TensorFlow \cite{environmental_setup_TF}. The decoder network employs deconvolution layers to produce the depth map by pixel-by-pixel estimation that matches the input size, while the encoder network utilizes convolution and pooling layers to acquire depth data. In the encoding and decoding stages, downsampling and upsampling are performed, respectively.

\begin{figure*}[h, align=\flushleft, width=\textwidth]
\centering
\includegraphics[width=0.7\textwidth]{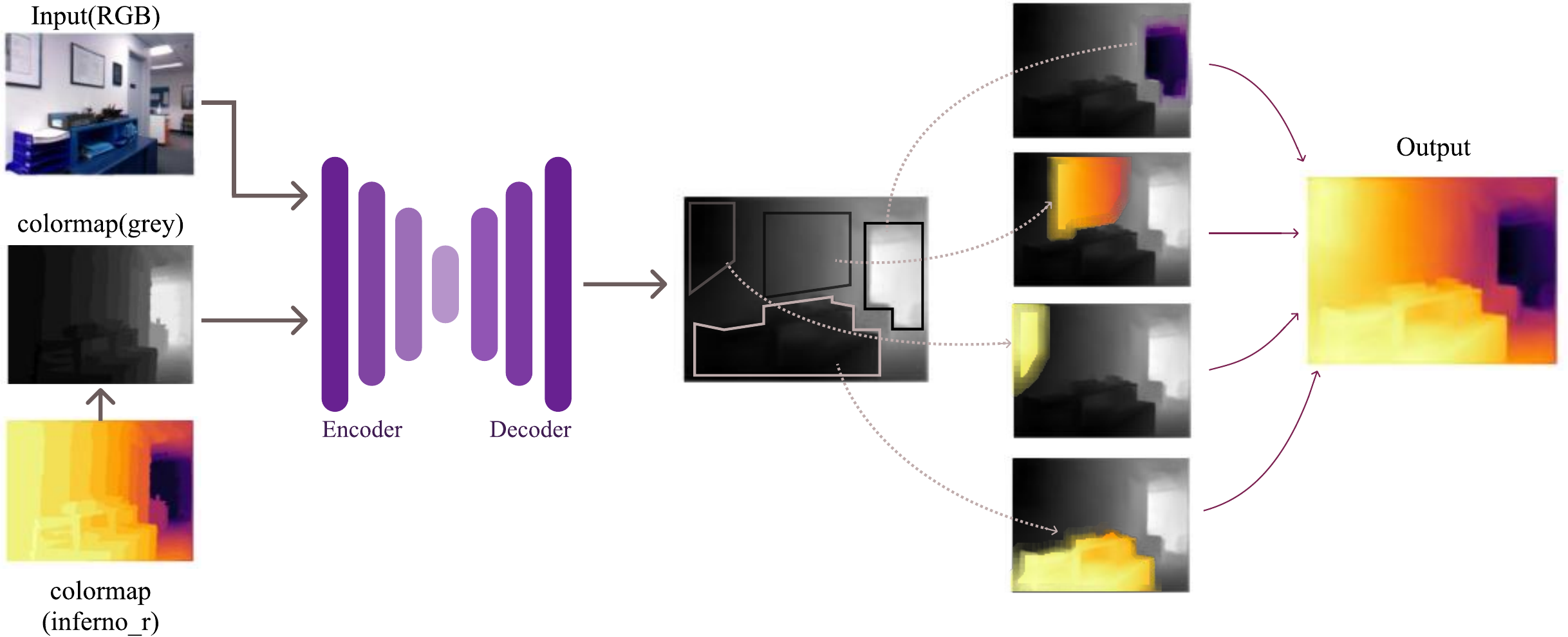}
\caption{The process of generating a color map by mapping depth information from a grayscale image through an encoder-decoder network.}
\label{fig}
\end{figure*}

\subsubsection{Encoder}
In the encoder section, we used the Inception-ResNet-v2 (IRv2) \cite{networkarchitecture_encoder_IRv2} architecture as a pre-trained model. Down-sampling is performed through the truncated layers of this pre-trained model. By passing the images through these layers, we will extract more accurate features from the images.
In Inception-Resnet-v2 architecture, three categories of inception blocks exist (Inception Resnet Block A, Inception Resnet Block B, Inception Resnet Block C), two types of reduction blocks (Reduction Block A and B), as well as a Stem Block and an Average Pooling layer. 
The inception blocks are repeated multiple times to form the full-layer architecture.

\begin{figure*}[h, align=\flushleft, width=\textwidth]
\centering
\includegraphics[width=\textwidth]{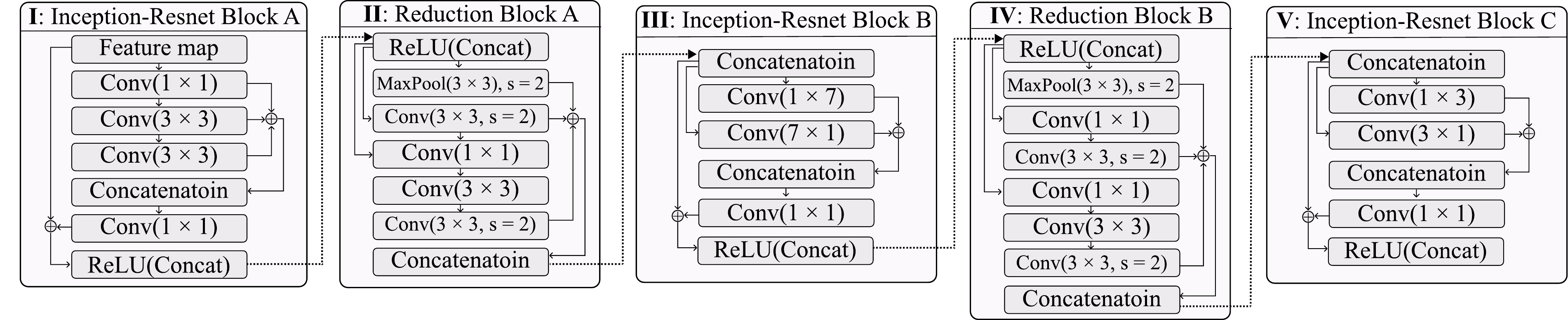}
\caption{Schematic Representation of Inception-Resnet A (IR-A), Reduction A (R-A), Inception-Resnet B (IR-B), Reduction B (R-B), and Inception-Resnet C (IR-C) Block}
\label{fig_algo_block}
\end{figure*}

\textbf{IR-A}. The Inception-ResNet-A block is a crucial component of the Inception-ResNet-v2 architecture. It is designed to capture multi-scale features and reduce the vanishing gradient problem. This block typically comprises multiple parallel branches, each utilizing different convolutional kernel sizes, including Conv(1x1), Conv(3x3), and Conv(5x5). To optimize parameter efficiency, a 1×1 convolutional filter precedes the 3×3 filter, and the 5×5 filter is factorized into two 3×3 filters. Residual connections within the block facilitate improved gradient flow during both forward and backward passes, addressing the vanishing gradient issue. The activation functions, typically ReLUs, are applied after each convolution to stabilize and activate the features. The structure of the Inception-ResNet-A block is represented in Figure \ref{fig_algo_block} (I).

\textbf{R-A}. The Reduction-A block reduces spatial dimensions and parameters using convolutional layers with strides greater than 1 or pooling operations. It is crucial for compressing feature maps while retaining essential information, improving computational efficiency, and generating more compact feature representations. The simplified visualization of the block is outlined in Figure \ref{fig_algo_block} (II).
\textbf{IR-B}. Inception-Resnet-B is another convolution block represented in Figure \ref{fig_algo_block} (III), consisting of a 7x7 convolution filter where the filter is factorized into 1x7 and 7x1, two asymmetric convolutions. By factorizing, the quantity of parameters is decreased, while the 1×1 convolution applied before the 7×7 convolution further limits the parameter count and enhances computational efficiency. 






\textbf{R-B}. Figure \ref{fig_algo_block} (IV) represents a simplified diagram of the Reduction-B block where the block reduces spatial dimensions while increasing channel depth, helping the Inception-ResNet-v2 model capture high-level features efficiently during downsampling.





\textbf{IR-C}. Inception-ResNet-C is another inception block in the Inception-ResNet-v2 architecture that uses a 3x3 convolution filter, preceded by a 1x1 convolution to reduce the number of parameters. Additionally, the 3x3 convolution is factorized into two asymmetric convolutions. The convolutions occur parallelly, and finally, all the outputs are concatenated. This allows the network to focus on capturing more abstract and higher-level data from the image. Here, Figure \ref{fig_algo_block} (V) illustrates the simplified representation of the block.\\

\begin{figure}[h]
\centering
  \includegraphics[width=.48\textwidth]{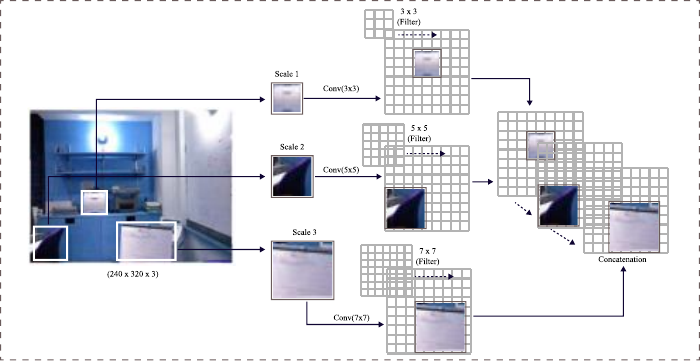}
\caption{Illustration of multi-scale feature extraction in Inception-ResNet v2, showcasing parallel branches with varying filter sizes and combining features across different scales.}
\label{fig:multi_scale_fe}
\end{figure}

The main advantage of using Inception-ResNet v2 \cite{networkarchitecture_encoder_IRv2} is, multi-scale feature extraction occurs through the use of Inception blocks, which are specifically designed to extract features at multiple scales simultaneously. Figure \ref{fig:multi_scale_fe} provides a clear illustration of how multi-scale feature extraction operates in Inception-ResNet v2. Each Inception module has parallel branches, where each branch processes the input features with different operations, such as 1x1, 3x3 and 7x7 convolutions. Next, Inception-ResNet modules introduce residual connections, where the output of the parallel branches is combined with the input and mitigate vanishing gradient problems. The architecture often uses factorized convolutions to reduce the computational cost of processing large filters. For example, a 3x3 convolution is factorized into 1D convolutions. Finally, the outputs of all branches in the Inception module are concatenated. Here, as the network progresses deeper, subsequent inception modules process the multi-scale features extracted earlier. This hierarchical processing allows the network to capture complex features at various scales.

\IncMargin{1em}
\begin{algorithm}[!ht]
\small
\caption{Inception-ResNet-v2 (IRv2) Encoder with Repeated Blocks}
\label{IRv2_encoder_referenced}
\SetAlgoLined
\SetKwInOut{Input}{Input}
\SetKwInOut{Output}{Output}

\Input{Feature map $x$}
\Output{Encoded feature map $IRv2\_Encoder(x)$}
\BlankLine

\textbf{Repeat IR\_A block 10 times:}\\
$i \leftarrow 1$\;
\While{$i \leq 10$}{
    \text{$x \leftarrow IR\_A(x)$\; [Figure \ref{fig_algo_block}(I)]}\\
    $i \leftarrow i + 1$\;
}

\textbf{Apply R\_A block:}\\
\text{$x \leftarrow R\_A(x)$\; [Figure \ref{fig_algo_block}(II)]}

\textbf{Repeat IR\_B block 5 times:}\\
$i \leftarrow 1$\;
\While{$i \leq 5$}{
    \text{$x \leftarrow IR\_B(x)$\; [Figure \ref{fig_algo_block}(III)]}\\
    $i \leftarrow i + 1$\;
}

\textbf{Apply R\_B block:}\\
\text{$x \leftarrow R\_B(x)$\; [{Figure \ref{fig_algo_block}(IV)}]}

\textbf{Repeat IR\_C block 10 times:}\\
$i \leftarrow 1$\;
\While{$i \leq 10$}{
    \text{$x \leftarrow IR\_C(x)$\;  [{Figure \ref{fig_algo_block}(V)}]}\\
    $i \leftarrow i + 1$\;
}

\Return $x$\;

\end{algorithm}
\DecMargin{1em}

\begin{figure*}[h, align=\flushleft, width=\textwidth]
\centering
\includegraphics[width=\textwidth]{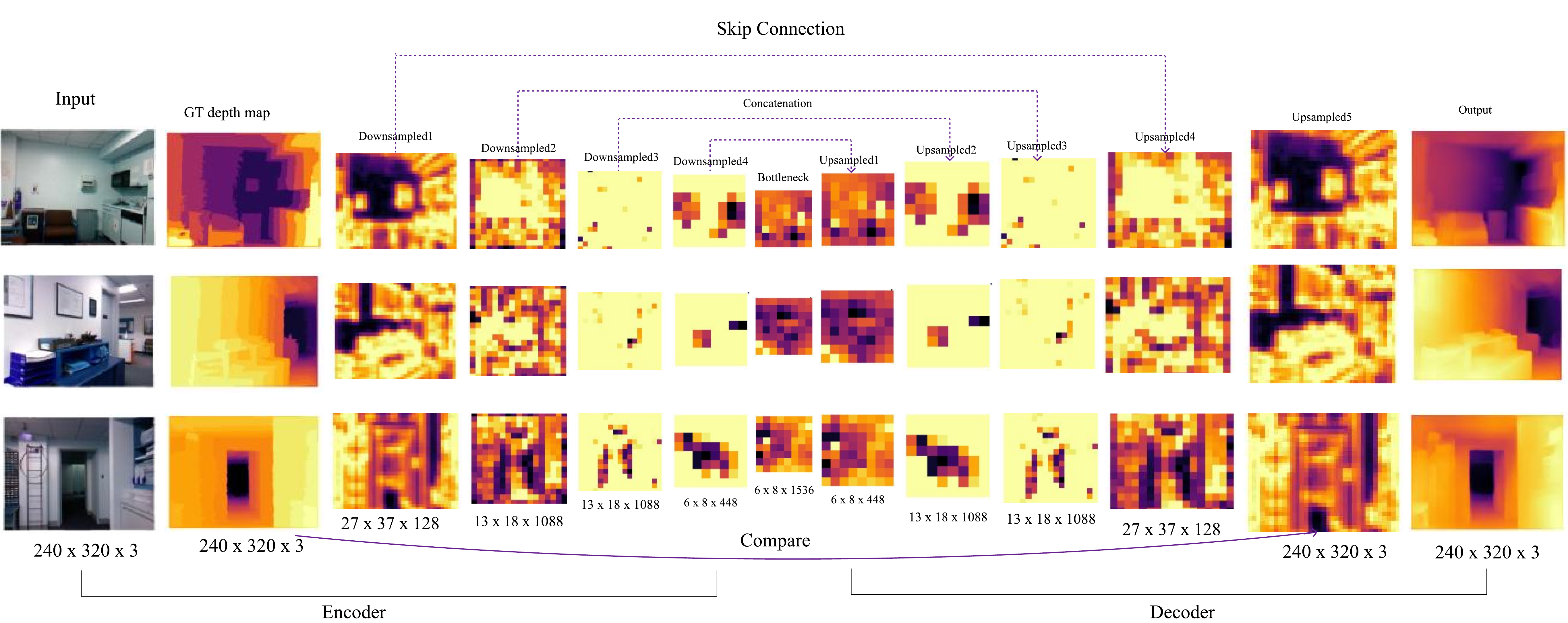}
\caption{Layer-by-layer feature map representation within the encoder-decoder network architecture, with Inception-ResNet-v2 (IRv2) as the encoder, designed for depth map generation.}
\label{fig}
\end{figure*}

\subsubsection{Decoder}
After downsampling the image in the encoder section, the resolution will become too low, losing some important features. As the number of features of the images is lower than the original size, it is necessary to reconstruct the image to increase the image's resolution before concatenation. The process of increasing the image size from a lower number of parameters to a higher number of parameters is called upsampling. This process will be started from the bottleneck situation of the architecture whether the bottleneck situation means the simplest form of the image with some specific important features. We use the LeakyRelu \cite{leakyrelu} activation function before starting upsampling from the bottleneck.
Here, LeakyRelu \cite{leakyrelu} is used to avoid neuron inactivity, ensures gradient flow, preserves information, and leads to faster, more stable convergence in this stage. Then upsampling of the image is necessary to concatenate the upsampled image with its corresponding downsampled image through a skip connection Figure \ref{fig1}. A skip connection is a mechanism in neural networks that allows the input from one layer to concatenate straight to a subsequent layer, skipping one or more intermediate layers in the process, helping to preserve important information and reduce vanishing gradient problems. This concatenation is necessary because it helps recover the features that were lost from the image during the earlier stages of processing. Here, skip connections enable the decoder to retrieve high-resolution feature maps from the encoder, guaranteeing the preservation of these features in the final depth map.
In the final stage of the model, an activation function (sigmoid) is applied to the depth map output to ensure that the predicted values are appropriately scaled between 0 and 1, providing a normalized representation of the depth information.

\subsection{Output}
The trained model generates a depth map from an input RGB image, representing object distances in the scene. In a grayscale map, closer objects appear darker, while farther ones are lighter, or vice versa. In a color depth map, the Inferno\_r colormap (from Matplotlib \cite{matplotlib}) is used, transitioning from bright yellow for closer objects to dark purple for farther ones.  In the case of color map representation, there are two steps that always happen. Firstly, the model generates a grayscale image as a depth map. This means that each pixel’s color is reduced to a single-intensity value. After that, the grayscale values are then mapped to colors in the Inferno\_r colormap. As a result, low grayscale values (close to black in grayscale) would be mapped to bright yellow, and high grayscale values (close to white in grayscale) would be mapped to dark purple.

\subsection{Loss Function} \label{subsec:loss_func}
The discrepancy between $y$, the ground truth depth map, and $\hat{y}$, the anticipated depth map, is taken into account by a typical loss function for depth regression issues. The training pace and total depth estimation performance can be greatly impacted by various loss function factors. A wide range of loss function modifications are used to optimize the neural network Neural network for estimating depth. In our method, a composite loss function equation \eqref{loss_func} is used, which helps to increase the model's accuracy by tuning the weight. For training the model, loss L is defined between $y$ and $\hat{y}$ as the weighted sum of three loss functions.
\begin{equation}
    L(y, \hat{y}) = w_1 L_{\text{depth}}(y, \hat{y}) + w_2 L_{\text{grad}}(y, \hat{y}) + w_3 L_{\text{SSIM}}(y, \hat{y})
\label{loss_func}
\end{equation}
Here, $w_1$, $w_2$, and $w_3$ are the weights assigned to different losses. $L_{\text{depth}}$, $L_{\text{grad}}$, and $L_{\text{SSIM}}$ are depth, gradient, and SSIM loss terms, respectively.
\subsubsection{Depth Loss}
Here, the loss term $L_{\text{depth}}$ refers to point-wise loss, a typical loss function for all techniques based on deep learning. The pixel-by-pixel discrepancy between the anticipated depth map and the actual depth map is estimated by calculating the average of these absolute discrepancies throughout every pixel in the image.
\begin{equation}
        L_{\text{depth}}(y, \hat{y}) = \frac{1}{n} \sum_{p}^{n} \left| y_p - \hat{y}_p \right|
        \label{eqn:eq2}
\end{equation}
\subsubsection{Gradient Edge Loss}
The term gradient edge loss $L_{\text{grad}}$ is measured by calculating the mean absolute disparity between the real depth and the anticipated depth's vertical and horizontal gradients.
\begin{equation}
        L_{\text{grad}}(y, \hat{y}) = \frac{1}{n} \sum_{p}^{n} \left| g_x(y_p , \hat{y}_p) \right| + \left| g_y(y_p , \hat{y}_p) \right|
        \label{eqn:eq2}
\end{equation}
Here, $g_x$ and $g_y$ represent horizontal edges and vertical edges, respectively.
\subsubsection{Structural Similarity (SSIM) Loss}
Finally, the loss term $L_{\text{SSIM}}$ is accustomed to determining how well the structural features are retained when comparing the projected depth map to the ground truth depth map. The initial task of calculating the structural loss is to determine the SSIM index.

\begin{equation}
        L_{\text{SSIM}}(y, \hat{y}) = \frac{1 - \text{SSIM}(y, \hat{y})}{2}
        \label{eqn:eq2}
\end{equation}
\begin{equation}
    \text{SSIM}(y, \hat{y}) = \frac{(2 \mu_y \mu_{\hat{y}} + C_1)(2 \sigma_{y\hat{y}} + C_2)}{(\mu_y^2 + \mu_{\hat{y}}^2 + C_1)(\sigma_y^2 + \sigma_{\hat{y}}^2 + C_2)}
\end{equation}
Here, the means of $y$ and $\hat{y}$ are denoted by $\mu_y$ and $\mu_{\hat{y}}$, respectively. The variances are $\sigma_y^2$ and $\sigma_{\hat{y}}^2$, and the covariance between $y$ and $\hat{y}$ is $\sigma_{y\hat{y}}$. The constants used to stabilize the division are $C_1$ and $C_2$.\\

In the composite loss function with three weighted components \(w_1\), \(w_2\), and \(w_3\) (each ranging between 0 and 1), we initially set all weights to 1. Then, we iteratively adjusted the weights one at a time to optimize model performance by reducing \( w_1 \) while keeping \( w_2 \) and \( w_3 \) constant. If performance degraded compared to the initial setup, we restored \(w_1\) to its previous value and moved on to adjust \(w_2\), and so on for \(w_3\). Using this iterative cross-checking process, we fine-tuned the values of \(w_1\), \(w_2\), and \(w_3\) for optimal model performance. The following algorithm \ref{optimize_weights_algo}  provides a detailed explanation of the process used for selecting the optimal weights.

\IncMargin{1em}
\begin{algorithm}[!ht]
\small
\caption{Optimizing Weights \(w_1, w_2, w_3\)}
\label{optimize_weights_algo}
\SetAlgoLined
\SetKwInOut{Input}{Input}
\SetKwInOut{Output}{Output}

\Input{Initial weights \(w_1 = w_2 = w_3 = 1\), Validation set}
\Output{Optimized weights \(w_1, w_2, w_3\)}

\BlankLine

\For{$w \in \{w_1, w_2, w_3\}$}{
    \textbf{Fix other weights, decrement} $w$ \;
    \While{performance improves}{
        \textbf{Validate performance} \;
        \If{performance deteriorates}{
            Revert $w$ \;
        }
    }
}

Cross-check and refine the values of \(w_1, w_2, w_3\) to ensure optimal adjustments\;

Evaluate on validation set\;

\Return{Optimized weights \(w_1, w_2, w_3\)}

\end{algorithm}
\DecMargin{1em}

\IncMargin{1em}
\begin{algorithm}[!ht]
\small
\caption{Training the Depth Prediction Model on NYU Depth v2}
\label{train_model_algo_encoder_decoder}
\SetAlgoLined
\SetKwInOut{Input}{Input}
\SetKwInOut{Output}{Output}

\Input{Training data $D_{\text{train}}$ from NYU Depth v2, Learning rate $\eta$, Number of iterations $T$, Weights $w_1$, $w_2$, $w_3$ for loss terms}
\Output{Trained model $\theta$}
\BlankLine

\textbf{Initialize the model and optimizer:}\\
$\theta \leftarrow \text{Random Initialization}$ \;
$\text{Optimizer} \leftarrow \text{Adam}(\theta, \eta)$\;

$t \leftarrow 0$\;
\While{$t < T$}{
    
    \textbf{Load batch of training data:} $(x, y) \leftarrow \text{LoadBatch}(D_{\text{train}})$\;

    \textbf{Forward pass (IRv2 Encoder-Decoder):}\\
    $z_{\text{encoder}} \leftarrow \text{IRv2\_Encoder}(x)$\; 
    
    
    $\hat{y} \leftarrow \text{Decoder}(z_{\text{encoder}})$\;
    

    \textbf{Compute total loss $L_{\text{total}}$:}\\
    $L_{\text{depth}} \leftarrow \text{MeanAbsoluteError}(y, \hat{y})$\;
    $L_{\text{grad}} \leftarrow \text{GradientEdgeLoss}(y, \hat{y})$\;
    $L_{\text{SSIM}} \leftarrow \text{SSIM\_Loss}(y, \hat{y})$\;
    $L_{\text{total}} \leftarrow w_1 L_{\text{depth}} + w_2 L_{\text{grad}} + w_3 L_{\text{SSIM}}$\;

    \textbf{Backpropagation and update:}\\
    $\text{Optimizer.zero\_gradients()}$\;
    $L_{\text{total}}.\text{backward()}$\;
    $\text{Optimizer.step()}$\;

    $t \leftarrow t + 1$\;
}

\Return $\theta$\;  

\end{algorithm}
\DecMargin{1em}

\section{Implementation}
\subsection{Dataset}
We utilized the NYU Depth V2 \cite{dataset_NYUv2} dataset to assess our suggested model. This contains RGB-D photos captured using the Microsoft Kinect sensor. It includes almost 1,400 densely annotated indoor scenes from 464 different locations, each one containing depth information and matching RGB photos. The dataset covers a variety of room types, such as living rooms, kitchens, and offices, as well as detailed per-pixel object tagging. From the 120,000 available samples in the dataset, a random subset of 65,000 samples was designated for training, while 654 samples were reserved for testing. We applied data augmentation to all images in the training set by horizontally flipping them while maintaining a constant height of 240 pixels to improve model training. The model generated predictions at a resolution of 240 × 320 × 3, which is half the input size (480 × 640 × 3). For training, the input images are down-scaled to 240 × 320 × 3, while the ground truth depth maps remain at their original resolution. Down-sampling reduces the computational load and speeds up training by decreasing the number of pixels the model needs to process, making it more efficient. During testing, the predicted depth map for the entire test image is computed and then up-scaled by a factor of two to match the ground truth resolution.\\
We used two additional datasets with outdoor scenes for our experiments. The first one is the KITTI \cite{dataset_KITTI} Depth Estimation dataset, a subset of the KITTI dataset specifically designed for depth estimation tasks using monocular or stereo images. For this study, 7,281 image pairs were used for training, while 200 image pairs were allocated for testing and evaluation. Each pair consists of a single RGB image and its corresponding ground truth depth map. The second dataset is the Cityscapes \cite{dataset_cityscapes} dataset, basically designed for semantic segmentation tasks to evaluate the model's performance in this operation. From this dataset, a subset of 1,572 image pairs was selected for training, while 500 image pairs were used for testing and evaluation. During preprocessing, all images were resized to 240 × 320 × 3 to ensure consistency of the data for both training and testing.

\subsection{Environmental Setup}
The proposed model was implemented using TensorFlow \cite{environmental_setup_TF}. The model training was conducted on an NVIDIA T4 GPU setup (2 GPUs), each with 16 GB of memory. The Adam \cite{environmental_setup_Adam} optimizer is used with AMSGrad enabled, a variant of gradient descent that adapts the learning rate dynamically was employed during training. Setting 0.0001 as the initial learning rate, and training was performed over 15 epochs. The default momentum parameters for Adam were maintained, with $\beta_1 = 0.9$ and $\beta_2 = 0.999$. The complete training process took approximately 5 hours with a batch size of 16. To mitigate overfitting, data augmentation techniques were applied, including horizontal flipping and cropping the height of the images to a constant size of 320 pixels. We did the same for other pre-trained models including VGG19 \cite{environmental_setup_vgg19}, ResNet50 \cite{environmental_setup_resnet}, ResNet152 \cite{environmental_setup_resnet}, DenseNet169 \cite{environmental_setup_densenet}, and DenseNet201 \cite{environmental_setup_densenet}. These models varied in the number of parameters, and their performance was also varied with different error values and accuracy scores. 


\subsection{Evaluation Metrics}
The error metrics used for evaluating the model's performance are defined by the following equations, which provide a quantitative assessment.
\\
\begin{itemize}
    \item Root Mean Squared Error (RMSE):
    \begin{equation}
        \text{RMSE} = \sqrt{\frac{1}{N} \sum_{i=1}^{N} (y_i - \hat{y}_i)^2}
        \label{eqn:eq2}
    \end{equation}

    \item Log10 Error:
    \begin{equation}
        \text{Log10} = \frac{1}{N} \sum_{i=1}^{N} \left| \log_{10}(y_i) - \log_{10}(\hat{y}_i) \right|
        \label{eqn:log10_error}
    \end{equation}

    \item Absolute Relative Error (ARE):
    \begin{equation}
        \text{ARE} = \frac{1}{N} \sum_{i=1}^{N} \left| \frac{y_i - \hat{y}_i}{y_i} \right|
        \label{eqn:eq3}
    \end{equation}

    \item Accuracy:
    Three different kinds of threshold values are used to evaluate the depth map's accuracy. A threshold in depth map prediction typically refers to a set of pre-defined limits within which the estimated depth values are considered to be accurate. The thresholds provide a straightforward way to evaluate the accuracy of depth predictions at different levels of precision.
    \begin{equation}
        \begin{split}
            \delta_i &= \frac{1}{N} \sum_{i=1}^{N} \left( \max\left( \frac{y_i}{\hat{y}_i}, \frac{\hat{y}_i}{y_i} \right) < \text{th}_i \right), \\
            &\text{where} \quad \text{th}_i \in \{1.25, 1.25^2, 1.25^3\}
        \end{split}
    \end{equation}

    where $\hat{y_i}$ is a pixel in the expected depth image $\hat{y}$ and $y_i$ is a pixel in depth image y. The total number of pixels for each depth picture is N, and $\delta$ represents the accuracy corresponding to the respective threshold values.
\end{itemize}

\section{Results}

\subsection{Comparison with State-of-the-Art Architectures across Various Datasets}

\begin{figure*}[ht, align=\flushleft, width=\textwidth]
\centering
\includegraphics[width=\textwidth]{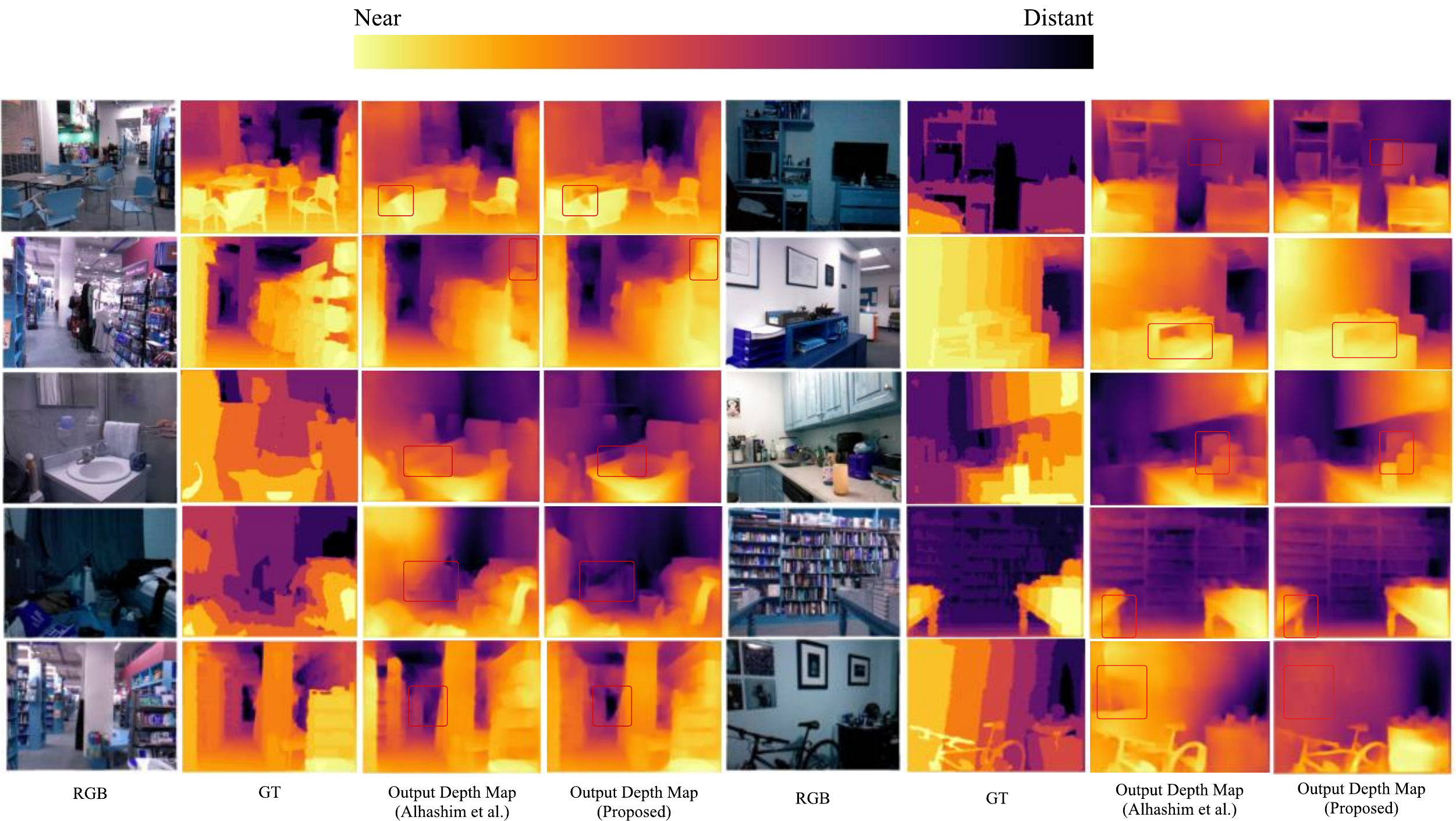}
\caption{Monocular depth estimation on the NYU-Depth v2 dataset using various encoder-decoder architectures, with output comparison between the method of Alhashim et al. \cite{alhashim_dense169} and the proposed model.}
\label{result_output}
\end{figure*}

\begin{figure*}[ht, align=\flushleft, width=\textwidth]
\centering
\includegraphics[width=\textwidth]{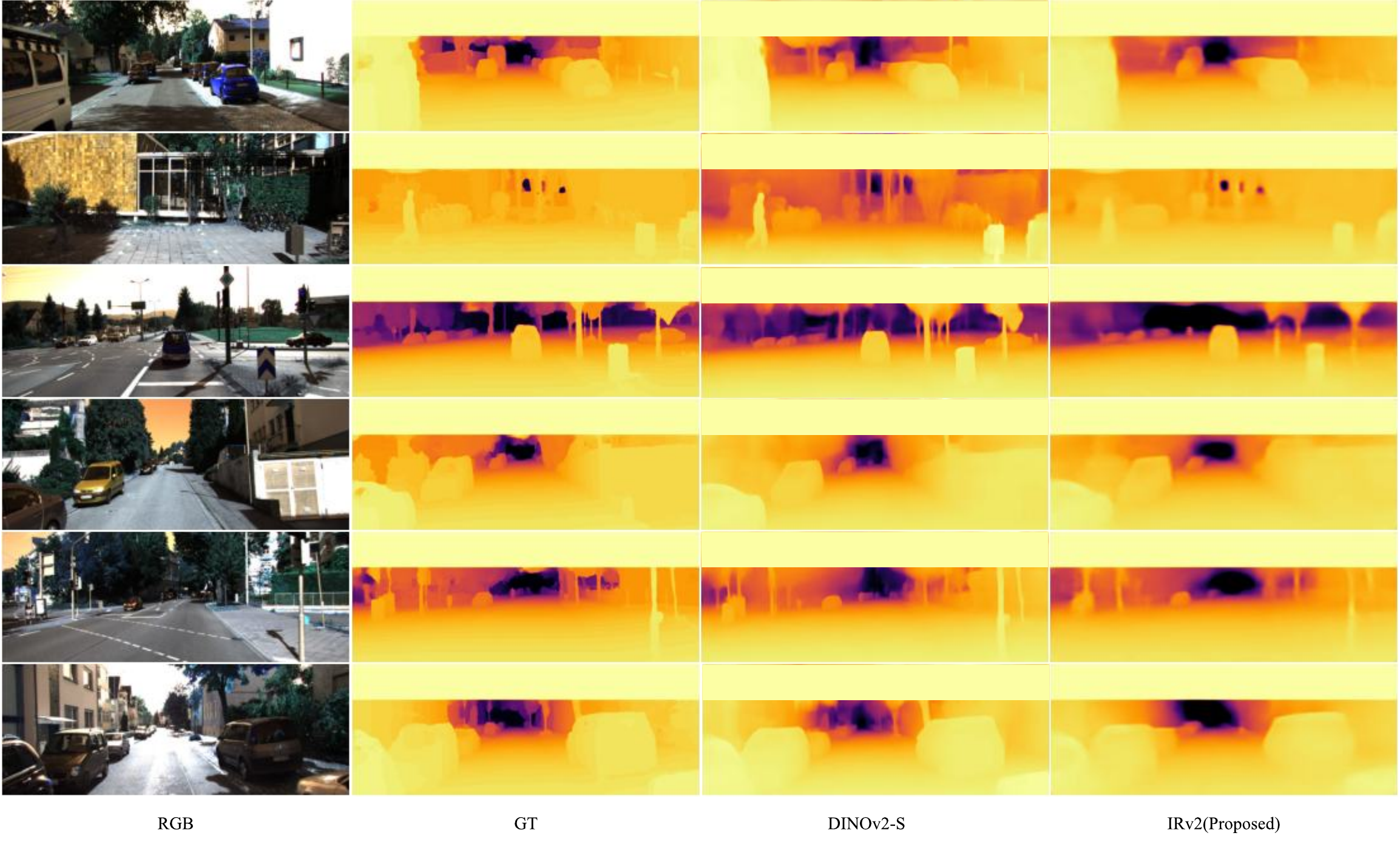}
\caption{Monocular depth estimation on the KITTI \cite{dataset_KITTI} dataset using two different architectures. From left to right: DINOv2-S \cite{oquab2023dinov2}, and Inception-ResNet v2 \cite{networkarchitecture_encoder_IRv2}.
}
\label{result_output_kitti}
\end{figure*}

\begin{figure}[htbp]
\begin{minipage}[t]{.49\textwidth} 
    \raggedleft
    \includegraphics[width=\textwidth]{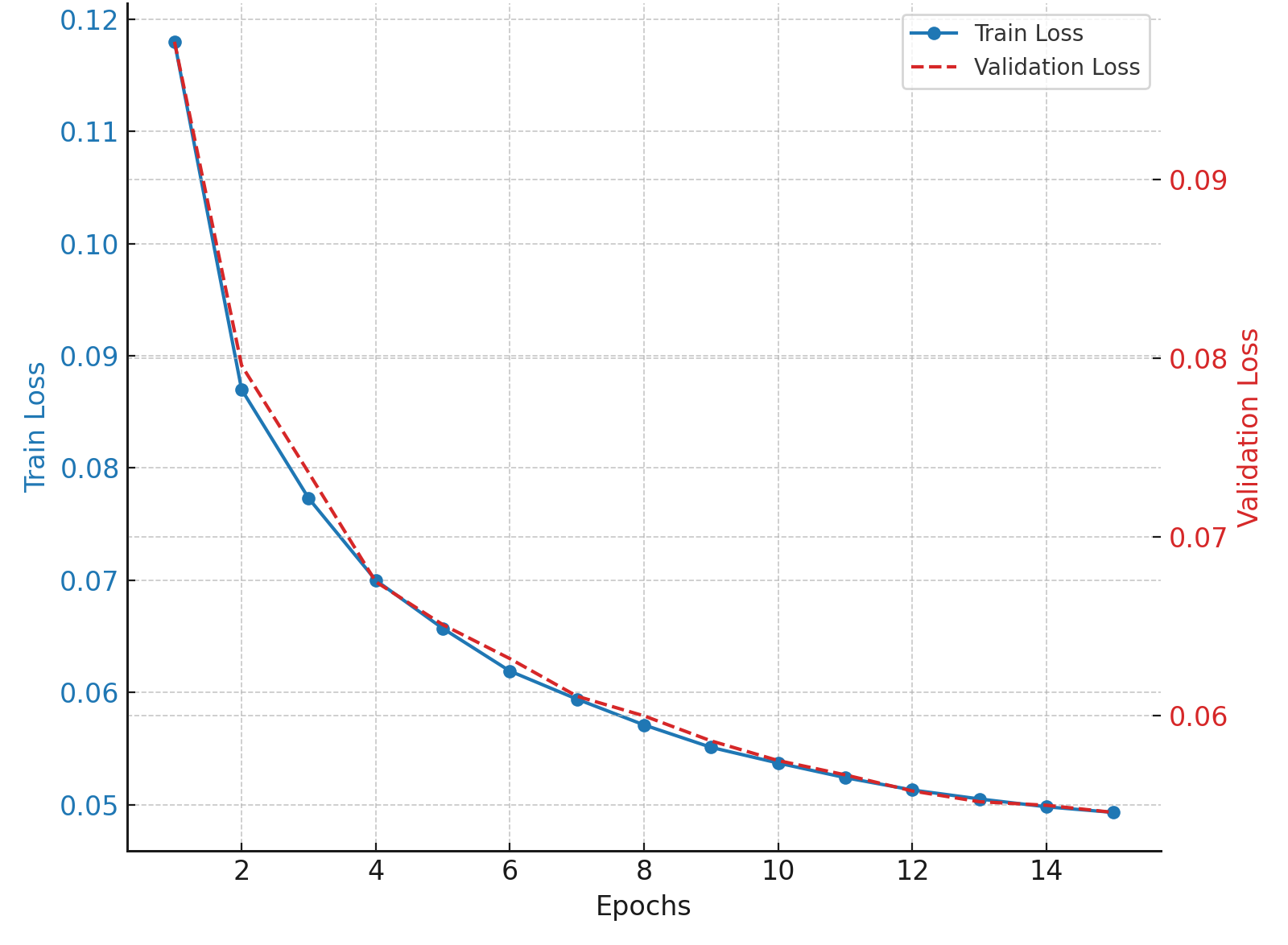}
    \caption{\textbf{Train and Validation Loss over Epochs.} This graph shows the decrease in both train loss (blue, solid line) and validation loss (red, dashed line) across 15 epochs on the NYU-
    Depth v2 dataset using IRv2 \cite{networkarchitecture_encoder_IRv2}.}
    \label{err_acc_fig}
\end{minipage}
\end{figure}

\begin{figure*}[ht, align=\flushleft, width=\textwidth]
\centering
\includegraphics[width=\textwidth]{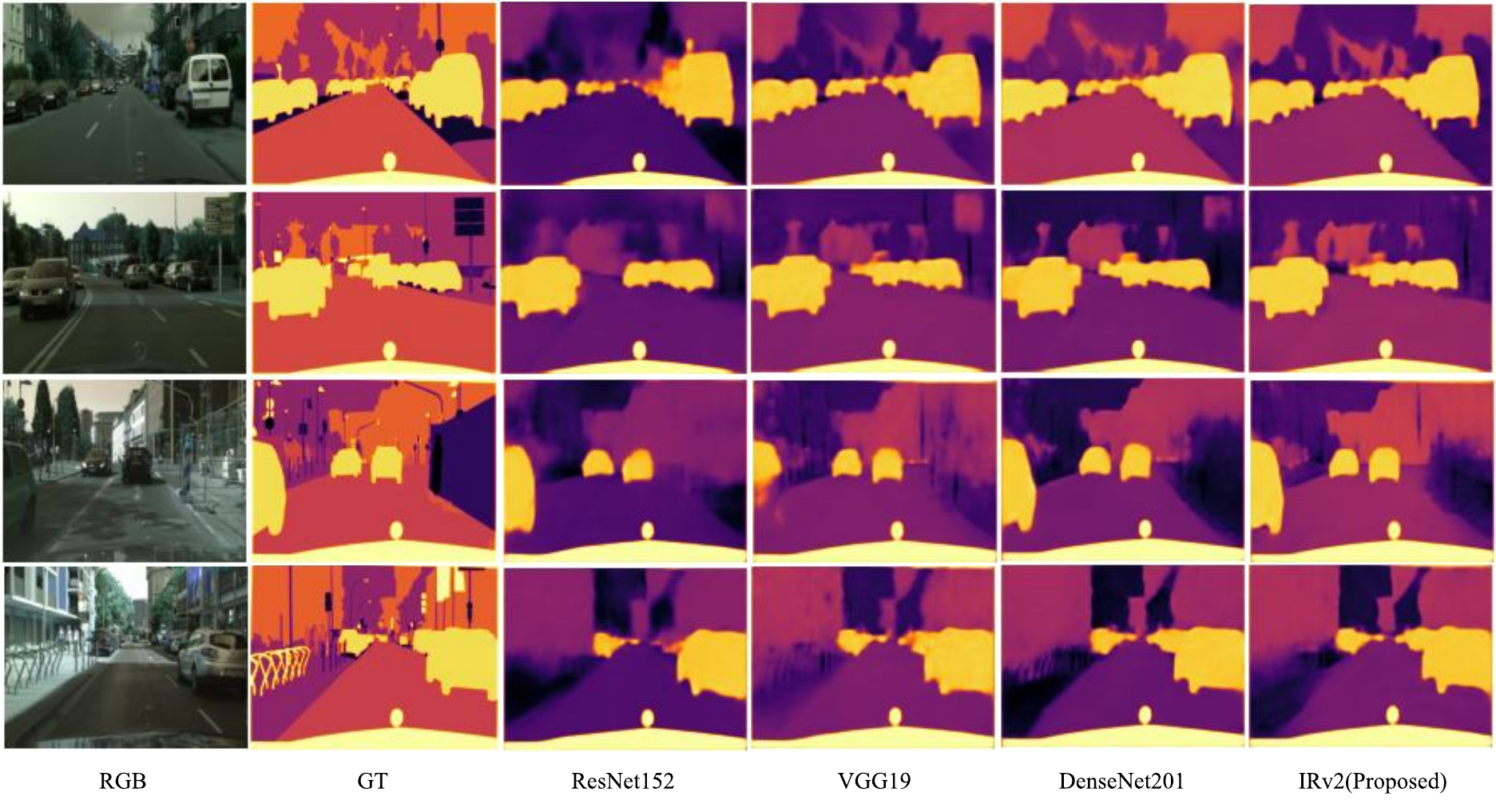}
\caption{Monocular depth estimation on the Cityscapes \cite{dataset_cityscapes} dataset using various encoder-decoder architectures. From left to right: ResNet152 \cite{environmental_setup_resnet}, VGG19 \cite{environmental_setup_vgg19}, DenseNet201 \cite{environmental_setup_densenet}, and Inception-ResNet v2 \cite{networkarchitecture_encoder_IRv2}.}
\label{result_output_cityscapes}
\end{figure*}

\begin{figure*}[h, align=\flushleft, width=\textwidth]
\centering
\includegraphics[width=\textwidth]{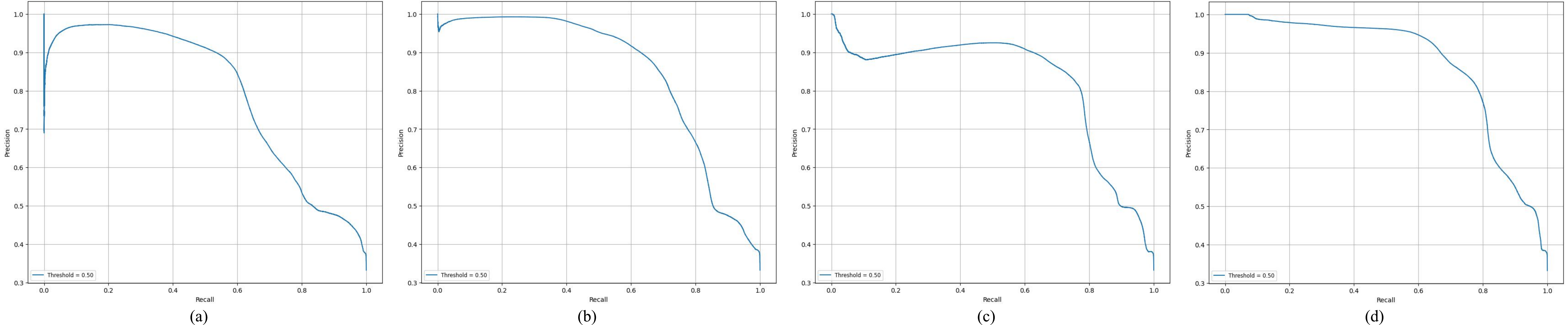}
\caption{Precision-Recall Curves for Different Encoder-Decoder Architectures on the Cityscapes \cite{dataset_cityscapes} dataset. The plots display the precision (y-axis) versus recall (x-axis) for various architectures: ResNet152 (a), VGG19 (b), DenseNet201 (c), and Inception-ResNet v2 (d).}
\label{prec_recall_curve}
\end{figure*}

\begin{figure*}[h, align=\flushleft, width=\textwidth]
\centering
\includegraphics[width=\textwidth]{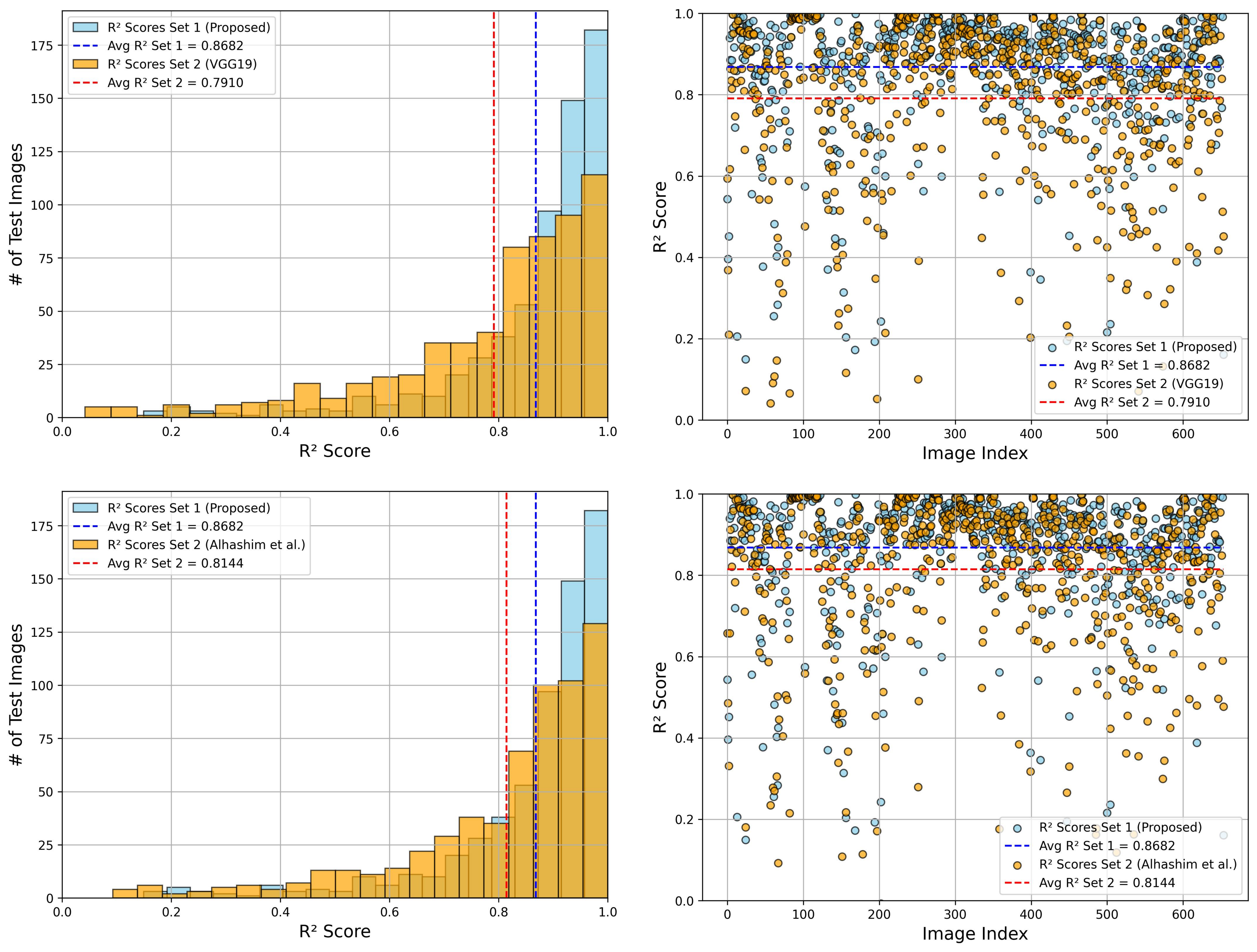}
\caption{\textbf{R\textsuperscript{2} Scores for Monocular Depth Estimation Across Test Images for Two Models on the NYU Depth v2.} The left plot shows the distribution of R\textsuperscript{2} values (x-axis) with frequency (y-axis), while the right plot displays individual R\textsuperscript{2} scores (y-axis) against image index (x-axis).}
\label{fig5}
\end{figure*}

\begin{table}[htbp]
\centering
\caption{Efficiency analysis of various encoder-decoder methods for monocular depth estimation on the NYU-Depth v2 \cite{dataset_NYUv2} dataset.}
\label{table:segmentation_comparison}
\renewcommand{\arraystretch}{1.5}
\footnotesize
\begin{tabular}{l p{6.0em} p{6.0em}}
\hline
\textbf{Method} & \textbf{Training Time (s)} & \textbf{Testing Time (s/sample)} \\ 
\hline
Enc-Dec-DenseNet201         & 18400 & 0.019 \\ 
Enc-Dec-ResNet152           & 15800 & 0.015 \\ 
Enc-Dec-VGG19               & 20115 & 0.019 \\ 
\textbf{Enc-Dec-IRv2 (Proposed)} & 17550 & 0.018 \\ 
\hline
\end{tabular}
\end{table}

\begin{table*}[htbp, align=\flushleft, width=\textwidth]
\caption{Performance comparison of various encoder-decoder methods for depth estimation on the NYU-Depth v2 dataset, where each method uses a different pre-trained encoder. All methods were implemented by the authors, with the proposed IRv2 model demonstrating superior performance across both error metrics and accuracy thresholds ($\delta < 1.25$, $\delta < 1.25^2$, $\delta < 1.25^3$).}
\label{table:comparison_flops}
\centering
\renewcommand{\arraystretch}{1.55}
\footnotesize	
\begin{tabular*}{\textwidth}{@{\extracolsep{\fill}} l c c c c c c c c }
\hline
\textbf{Method} & \textbf{Params} & \textbf{FLOPs} & \multicolumn{3}{c}{\textbf{Error Metrics} $\downarrow$} & \multicolumn{3}{c}{\textbf{Accuracy} $\uparrow$} \\ 
\cline{4-6} \cline{7-9}
& & & \textbf{ARE} & \textbf{RMSE} & \textbf{Log10} & \textbf{$\delta < 1.25$} & \textbf{$\delta < 1.25^2$} & \textbf{$\delta < 1.25^3$} \\ 
\hline
Enc-Dec-DenseNet201              & 17.23 M & 1.6e10 & 0.101 & 0.320 & 0.051 & 0.831 & 0.956 & 0.979 \\ 
Enc-Dec-ResNet152                & 15.54 M & 7.6e09 & 0.117 & 0.364 & 0.058 & 0.796 & 0.942 & 0.970 \\ 
Enc-Dec-VGG19                    & 22.37 M & 1.3e12 & 0.110 & 0.324 & 0.054 & 0.820 & 0.950 & 0.976 \\ 
\textbf{Enc-Dec-IRv2 (Proposed)} & \text{31.15 M} & 1.1e11 & \textbf{0.064} & \textbf{0.228} & \textbf{0.032} & \textbf{0.893} & \textbf{0.967} & \textbf{0.985} \\ 
\hline
\end{tabular*}
\end{table*}

\begin{table*}[htbp, align=\flushleft, width=\textwidth]
\caption{Performance comparison of various existing methods for depth estimation using error metrics and accuracy at different thresholds ($\delta < 1.25$, $\delta < 1.25^2$, $\delta < 1.25^3$) on the NYU-Depth v2 dataset, where the proposed model outperforms the others.}
\label{table:comparison2}
\centering
\renewcommand{\arraystretch}{1.55}
\footnotesize	
\begin{tabular*}{\textwidth}{@{\extracolsep{\fill}} l c c c c c c }
\hline
\textbf{Method} & \multicolumn{3}{c}{\textbf{Error Metrics} $\downarrow$} & \multicolumn{3}{c}{\textbf{Accuracy} $\uparrow$} \\ 
\cline{2-4} \cline{5-7}
& \textbf{ARE} & \textbf{RMSE} & \textbf{Log10} & \textbf{$\delta < 1.25$} & \textbf{$\delta < 1.25^2$} & \textbf{$\delta < 1.25^3$} \\ 
\hline
Rudolph et al. \cite{Rudolph_DM}    & 0.138 & 0.501 & 0.058 & 0.823 & 0.961 & 0.990 \\ 
Basak et al. \cite{Basak_ed_TL}     & 0.103 & 0.388 & --    & 0.892 & \textbf{0.978} & \textbf{0.995} \\ 
Alhashim et al. \cite{alhashim_dense169} & 0.123 & 0.465 & 0.053 & 0.846 & 0.974 & 0.994 \\ 
Lee et al. \cite{CNN5}              & 0.131 & 0.538 & --    & 0.837 & 0.971 & 0.994 \\ 
Xu et al. \cite{CNN6}               & 0.125 & 0.593 & 0.057 & 0.806 & 0.952 & 0.986 \\ 
Jung et al. \cite{GAN1}             & 0.134 & 0.527 & --    & 0.822 & 0.971 & 0.993 \\ 
Li et al. (VGG16) \cite{CNN2}       & 0.152 & 0.611 & 0.064 & 0.789 & 0.955 & 0.988 \\ 
Li et al. (VGG19) \cite{CNN2}       & 0.146 & 0.617 & 0.063 & 0.795 & 0.958 & 0.991 \\ 
Li et al. (ResNet50) \cite{CNN2}    & 0.143 & 0.635 & 0.063 & 0.788 & 0.958 & 0.991 \\ 
\textbf{Enc-Dec-IRv2 (Proposed)} & \textbf{0.064} & \textbf{0.228} & \textbf{0.032} & \textbf{0.893} & 0.967 & 0.985 \\ 
\hline
\end{tabular*}
\end{table*}

\begin{table*}[htbp, align=\flushleft, width=\textwidth]
\caption{Performance comparison of the proposed method against various Vision Transformer (ViT) models for depth estimation. The comparison is based on inference time at different input dimensions on the KITTI \cite{dataset_KITTI} dataset using 2x NVIDIA T4 GPUs.}

\label{table:kitti_table_inference}
\centering
\renewcommand{\arraystretch}{1.55}
\footnotesize
\begin{tabular*}{\textwidth}{@{\extracolsep{\fill}} l c c c }
\hline
\textbf{Method} & \multicolumn{3}{c}{\textbf{Inference Time (s/sample)} $\downarrow$} \\ 
\cline{2-4}
& \textbf{Input Dim (224 x 224)} & \textbf{Input Dim (320 x 240)} & \textbf{Input Dim (1242 x 375)} \\ 
\hline
DINOv2 (ViT-G) \cite{oquab2023dinov2} & 3.651 & 5.652 & 76.19 \\ 
DINOv2 (ViT-L) \cite{oquab2023dinov2} & 1.296 & 2.177 & 31.32 \\ 
DINOv2 (ViT-B) \cite{oquab2023dinov2} & 0.667 & 1.129 & 14.14 \\ 
DINOv2 (ViT-S) \cite{oquab2023dinov2} & 0.440 & 0.781 & 8.959 \\ 
\textbf{Enc-Dec-IRv2 (Proposed)}      & -- & \textbf{0.019} & -- \\ 
\hline
\end{tabular*}
\end{table*}

\begin{table*}[htbp, align=\flushleft, width=\textwidth]
\caption{Performance comparison of the proposed method against various Vision Transformer (ViT) models for depth estimation. The comparison is based on error metrics and accuracy at different thresholds ($\delta < 1.25$, $\delta < 1.25^2$, $\delta < 1.25^3$) on the KITTI \cite{dataset_KITTI} dataset.}

\label{table:kitti_table_accuracy}
\centering
\renewcommand{\arraystretch}{1.55}
\footnotesize
\begin{tabular*}{\textwidth}{@{\extracolsep{\fill}} l c c c c c c c }
\hline
\textbf{Method} & \textbf{Params} & \multicolumn{3}{c}{\textbf{Error Metrics} $\downarrow$} & \multicolumn{3}{c}{\textbf{Accuracy} $\uparrow$} \\ 
\cline{3-5} \cline{6-8}
& & \textbf{ARE} & \textbf{RMSE} & \textbf{Log10} & \textbf{$\delta < 1.25$} & \textbf{$\delta < 1.25^2$} & \textbf{$\delta < 1.25^3$} \\ 
\hline
DINOv2 (ViT-G) \cite{oquab2023dinov2} & 1196.05 M & \textbf{0.065} & \textbf{2.112} & \textbf{0.038} & \textbf{0.968} & \textbf{0.997} & \textbf{0.999} \\ 
DINOv2 (ViT-L) \cite{oquab2023dinov2} & 337.10 M  & 0.082 & 2.788 & 0.062    & 0.952 & 0.993 & 0.998 \\ 
DINOv2 (ViT-B) \cite{oquab2023dinov2} & 109.70 M   & 0.089 & 2.965 & 0.077 & 0.937 & 0.990 & 0.995 \\ 
DINOv2 (ViT-S) \cite{oquab2023dinov2} & 35.48 M   & 0.101 & 3.208 & 0.094    & 0.923 & 0.986 & 0.993 \\ 
\textbf{Enc-Dec-IRv2 (Proposed)}      &  31.15 M   & 0.124 & 3.720 & 0.110 & 0.899 & 0.972 & 0.989 \\ 
\hline
\end{tabular*}
\end{table*}

\begin{table*}[htbp, align=\flushleft, width=\textwidth]
\centering
\caption{Performance comparison of various encoder-decoder methods for segmentation tasks on the Cityscapes \cite{dataset_cityscapes} dataset. The table highlights training and testing times along with segmentation accuracy metrics such as Mean IoU, Precision, Recall, F1-Score, and Accuracy.}
\label{table:segmentation_comparison_table}
\renewcommand{\arraystretch}{1.5}
\footnotesize	
\begin{tabular*}{\textwidth}{@{\extracolsep{\fill}} l c p{5.9em} c c c c c c }
\hline
\textbf{Method} & \textbf{Training Time (s)} & \textbf{Testing Time (s/sample)} & \textbf{Mean IoU} & \textbf{Precision} & \textbf{Recall} & \textbf{F1-Score} & \textbf{Accuracy (\%)} \\ 
\hline
Enc-Dec-DenseNet201              & 3314 & 0.018 & 0.842 & 0.838 & 0.763 & 0.781 & 90.2 \\ 
Enc-Dec-ResNet152                & 3020 & 0.014 & 0.785 & 0.893 & 0.602 & 0.671 & 84.2 \\ 
Enc-Dec-VGG19                    & 3240 & 0.019 & 0.818 & \textbf{0.899} & 0.645 & 0.724 & 89.3 \\ 
\textbf{Enc-Dec-IRv2 (Proposed)}      & 3263 & 0.017 & \textbf{0.854} & 0.848 & \textbf{0.812} & \textbf{0.818} & \textbf{91.5} \\ 
\hline
\end{tabular*}
\end{table*}

 From table \ref{table:comparison_flops}, the proposed encoder-decoder method with IRv2 achieves a $\delta$ $<1.25$ accuracy of 0.893, which is significantly higher than most competing methods. This high accuracy at a strict threshold ($\delta$ $<1.25$) indicates that the method can reliably estimate depth with high precision, even when objects in the image are complex and vary in distance. The table \ref{table:comparison_flops} also presents the computational complexity of the models measured in Floating Point Operations (FLOPs). The proposed model has a computational complexity of \(1.1 \times 10^{11}\) FLOPs, which is higher than DenseNet201 \cite{environmental_setup_densenet} and ResNet152 \cite{environmental_setup_resnet}. However, VGG19 \cite{environmental_setup_vgg19} achieves the highest computational cost with the highest FLOPs among all the compared models. In Figure \ref{result_output}, the outputs of two different models are presented. The marked areas for Alhashim et al. \cite{alhashim_dense169} and our model highlight the differences in depth prediction accuracy when compared to the ground truth depth map. From the marked area, it is evident that the complex objects or the complex portion of the objects are distinguished more precisely by our model. Besides, our model achieves the lowest ARE of 0.064, showing that it is more consistent in estimating depth values, with the lowest RMSE of 0.228, indicating that the predicted depth values align more closely with the true values. The model also excels in the Log10 error, scoring 0.032, showing superior performance in minimizing the magnitude of error in the logarithmic scale.
 
 Although some studies demonstrate improved accuracy under specialized scenarios, such as the work by Li et al. \cite{RADepth}, their model outperforms ours by leveraging the combination of multiscale boundary features and reflectance to enhance overall prediction accuracy.
 
During training, the model's loss gradually decreases until it converges with the validation loss. At this stage, training is terminated, resulting in an optimal loss value (0.1523) for our model by selecting the appropriate number of epochs. Figure \ref{err_acc_fig} illustrates the variations of training loss and validation loss after each epoch.

The histograms located in the top-left and bottom-left corners of Figure \ref{fig5} compare the R\textsuperscript{2} scores of our proposed model (depicted in blue) against the encoder-decoder model using VGG19 and the architecture proposed by Alhashim et al. \cite{alhashim_dense169} (shown in orange).  Additionally, the scatter plots on the right side of Figure \ref{fig5} display the R\textsuperscript{2} scores for individual test images. Our model significantly outperforms these alternatives, achieving an R\textsuperscript{2} score of 0.8682.\\

After that, the proposed model was evaluated on the outdoor KITTI dataset \cite{dataset_KITTI}. A comparative analysis was conducted between our proposed model, IRv2 \cite{networkarchitecture_encoder_IRv2}, and the Vision Transformer-based model, DINOv2 \cite{oquab2023dinov2}. As illustrated in Figure \ref{result_output_kitti}, our model demonstrates results that closely align with DINOv2-S, which is the most efficient and smallest variant among the DINOv2 models. The results were obtained using high-resolution input images (1242 × 375).

Table \ref{table:kitti_table_inference} presents a comparison of the inference times of the proposed model with different variants of DINOv2 \cite{oquab2023dinov2}. Inference time means the amount of time a model takes to process an image and produce a result. To evaluate inference time, images (RGB) of three different dimensions were used: 1242 × 375 (the raw image size from the dataset), 224 × 224 (the dimension used by the ViT model), and 320 × 240 (the dimension used by the proposed model). It was observed that resizing the input images significantly reduced inference time. For instance, the smallest variant of DINOv2 achieved the lowest inference time of 0.440 seconds with 224 × 224 image resizing. Similarly, for the 320 × 240 image size, the smallest DINOv2 variant had an inference time of 0.781 seconds, which was faster than other DINOv2 variants. However, when compared to our proposed model, the IRv2 significantly outperformed DINOv2-S in terms of efficiency, with inference times nearly 41 times faster than DINOv2-S and 23 times faster than the best performance of DINOv2 \cite{oquab2023dinov2}. Although the accuracy of our proposed model is slightly (2.6\%) lower than the small variant of Transformer-based models, it demonstrates remarkable efficiency in terms of inference time, surpassing DINOv2-S and other variants, as shown in Table \ref{table:kitti_table_inference} and \ref{table:kitti_table_accuracy}. This highlights the potential of our model for real-time applications where computational efficiency is critical.\\

The model was further trained on the Cityscapes \cite{dataset_cityscapes} dataset, which is primarily designed for semantic segmentation task. Statistical analyses were conducted to compare the performance of the proposed model with other state-of-the-art CNN-based encoder-decoder architectures. As shown in table \ref{table:segmentation_comparison_table}, the proposed model achieved the highest performance across multiple metrics, including mean Intersection over Union (mIoU), Recall, F1-Score, and Accuracy. The Precision-Recall curve presented in Figure \ref{prec_recall_curve} illustrates the performance of all evaluated models, highlighting that the proposed model outperforms others. Additionally, Figure \ref{result_output_cityscapes} provides a side-by-side comparison of the segmentation outputs, illustrating that the proposed model produces better-segmented results compared to the other models.

\subsection{Ablation Study}
\subsubsection{Layer-wise Feature Extraction with IRv2}
\begin{figure}[h]
\centering
  \includegraphics[width=.48\textwidth]{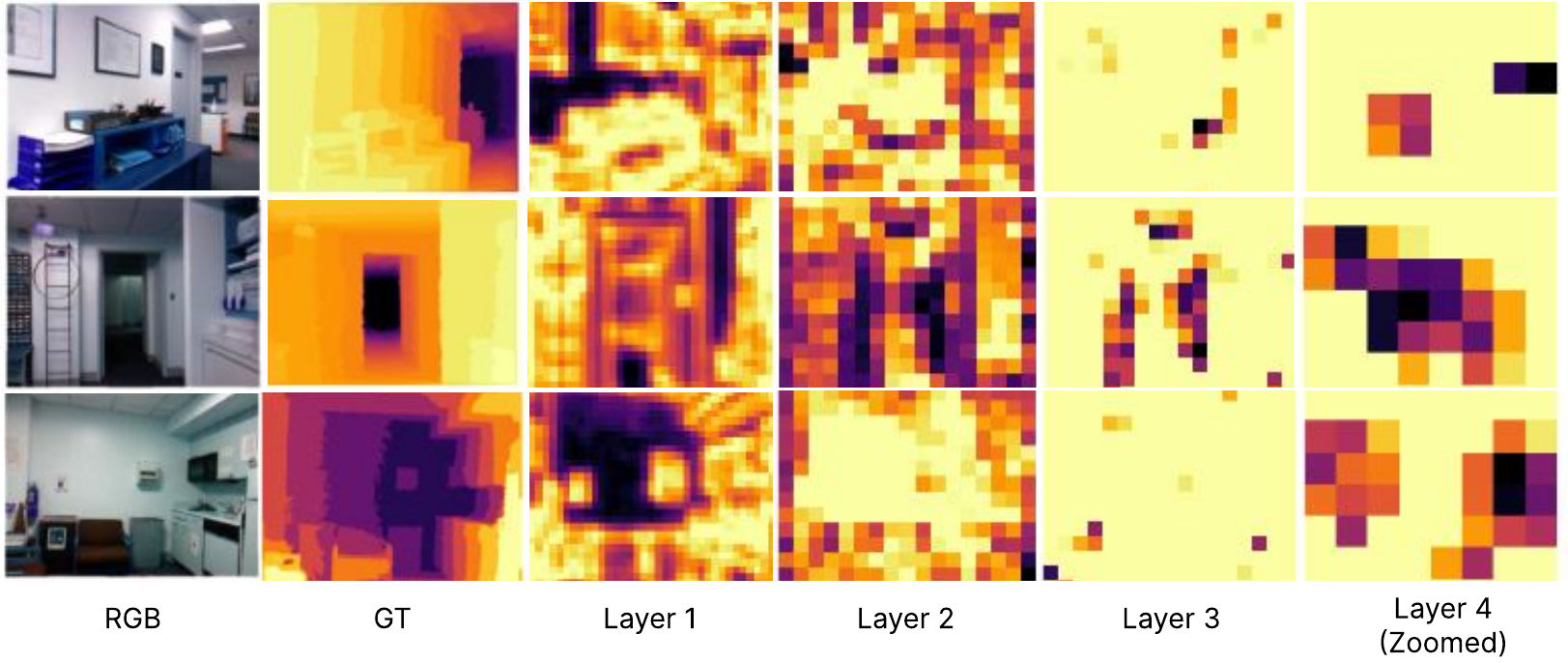}
\caption{Visual representation of some layers and the corresponding extracted features on the NYU-Depth v2 dataset using IRv2 \cite{networkarchitecture_encoder_IRv2}.}
\label{fig:layer_by_layer_presentation}
\end{figure}

Our proposed architecture uses IRv2 \cite{networkarchitecture_encoder_IRv2} layers as the encoder for feature extraction. This combines the efficient multi-branch architecture of the Inception network with the residual connections of ResNet. By leveraging these layers, the encoder efficiently extracts rich, high-level representations from the input RGB images and this feature extraction mechanism enhances the model’s ability to handle complex visual information. Figure \ref{fig:layer_by_layer_presentation} provides a visual representation of each layer along with its extracted features.

\subsubsection{Composite Loss Function}
\begin{figure*}[ht, align=\flushleft, width=\textwidth]
\centering
\includegraphics[width=\textwidth]{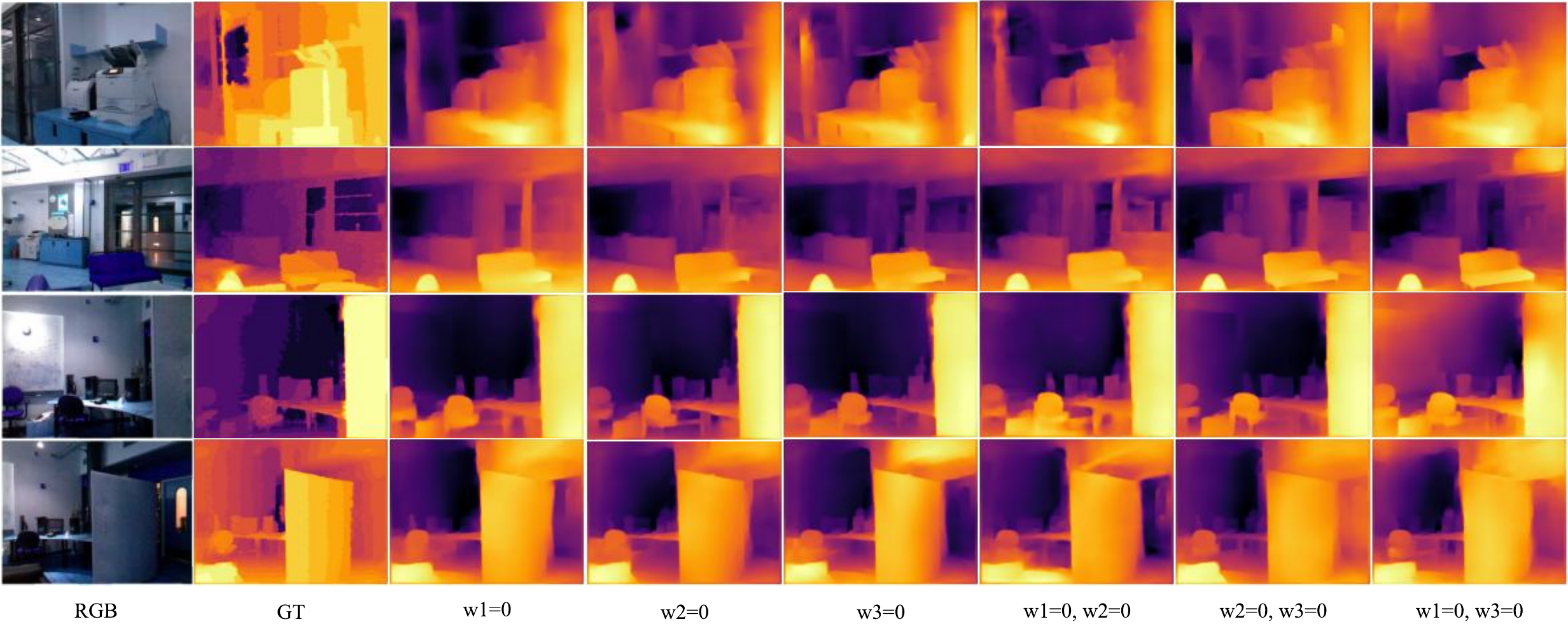}
\caption{The visualization of performance comparison on the NYU-Depth v2 dataset using IRv2 across various weight combinations in the composite loss function.}
\label{output_weight_combination}
\end{figure*}

The composite loss function consists of three components: Structural Similarity Loss (SSIM), Edge Loss, and Depth Loss, combined as a weighted sum where the corresponding weights \(w_1\), \(w_2\), and \(w_3\) with selected values (e.g., \ref{subsec:loss_func}, Loss Function) are applied to Depth, Edge, and SSIM losses, respectively. To evaluate the effect of each component, different combinations are considered by setting specific weights to zero. When \(w_1 = 0\), Depth Loss is excluded, and the loss function uses only Edge and SSIM losses. Similarly, \(w_2 = 0\) excludes Edge Loss, and \(w_3 = 0\) excludes SSIM Loss, retaining the remaining two losses. Additionally, we analyze cases with a single loss: \(w_1 = 0, w_2 = 0\) considers only SSIM Loss; \(w_1 = 0, w_3 = 0\) considers only Edge Loss; and \(w_2 = 0, w_3 = 0\) considers only Depth Loss. Finally, the results are presented for the case where all three types of loss are combined. Table \ref{table:weight_comparison} presents the performance of the model using different combinations of three different losses of the composite loss function and Figure \ref{output_weight_combination} illustrates the individual outputs corresponding to each combination.

\begin{table*}[htbp, align=\flushleft, width=\textwidth]
\caption{Performance comparison using error metrics and accuracy at different thresholds ($\delta < 1.25$, $\delta < 1.25^2$, $\delta < 1.25^3$) on the NYU-Depth v2 dataset using IRv2 across various weight combinations in the composite loss function.}
\label{table:weight_comparison}
\centering
\renewcommand{\arraystretch}{1.55}
\footnotesize
\begin{tabular*}{\textwidth}{@{\extracolsep{\fill}} l c c c c c c}
\hline
\textbf{Weight Combination} & \textbf{ARE} & \textbf{RMSE} & \textbf{Log10} & \textbf{$\delta < 1.25$} & \textbf{$\delta < 1.25^2$} & \textbf{$\delta < 1.25^3$} \\ 
\hline
\textbf{$w_1 = 0$}            & 0.067 & 0.230 & 0.036 & 0.889 & 0.964 & 0.982 \\ 
\textbf{$w_2 = 0$}            & 0.066 & 0.236 & 0.034 & 0.885 & 0.960 & 0.979 \\ 
\textbf{$w_3 = 0$}            & 0.069 & 0.229 & 0.038 & 0.876 & 0.962 & 0.980 \\ 
\textbf{$w_1 = 0, w_2 = 0$}   & 0.091 & 0.332 & 0.061 & 0.861 & 0.947 & 0.968 \\ 
\textbf{$w_2 = 0, w_3 = 0$}   & 0.088 & 0.339 & 0.055 & 0.864 & 0.949 & 0.970 \\ 
\textbf{$w_1 = 0, w_3 = 0$}   & 0.097 & 0.401 & 0.077 & 0.845 & 0.940 & 0.966 \\ 
\textbf{$w_1 \neq 0, w_2 \neq 0, w_3 \neq 0$} & \textbf{0.064} & \textbf{0.228} & \textbf{0.032} & \textbf{0.893} & \textbf{0.967} & \textbf{0.985}\\
\hline
\end{tabular*}
\end{table*}

\subsubsection{Depth Prediction under Degradations}
\begin{figure}[h]
\centering
  \includegraphics[width=.48\textwidth]{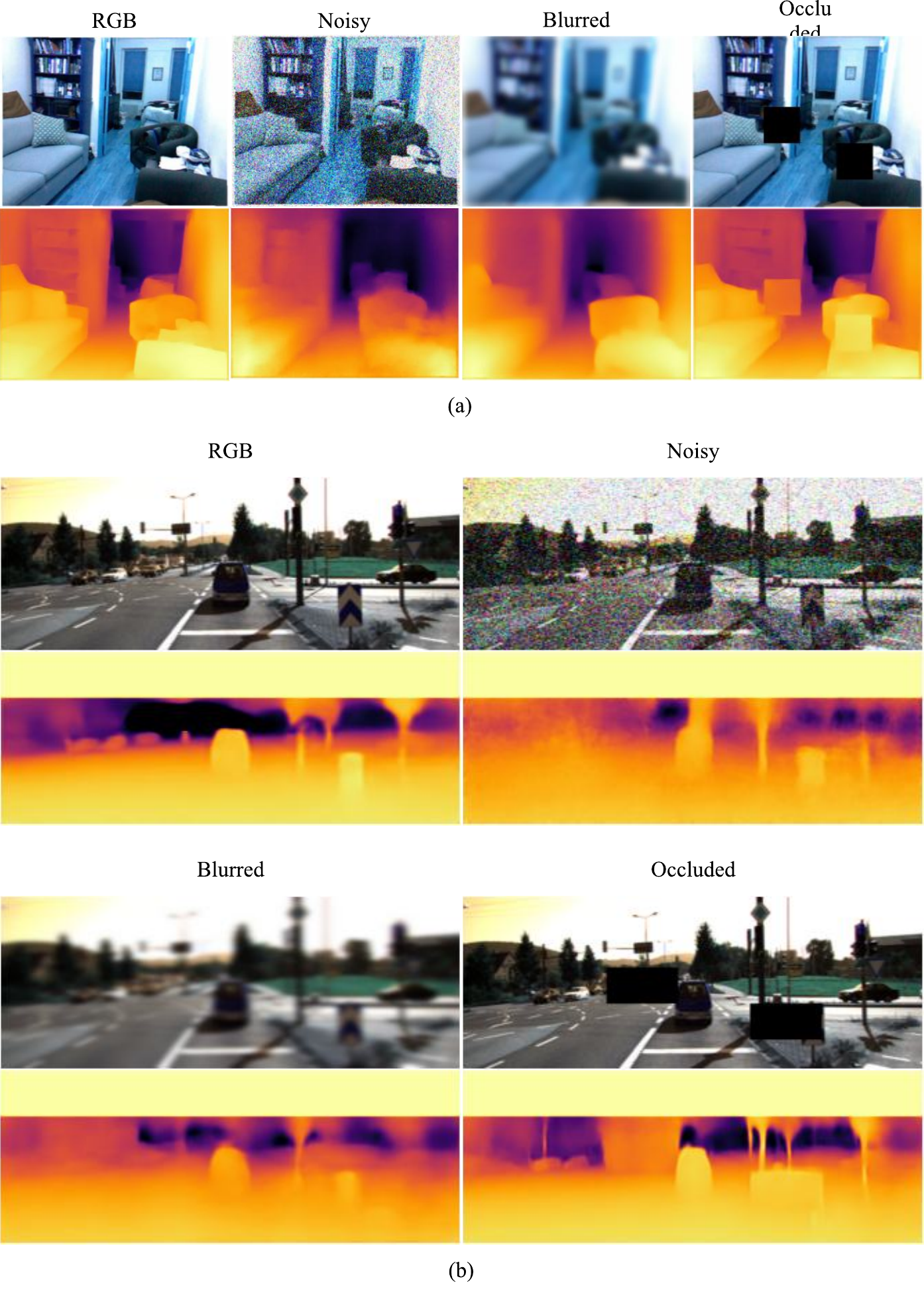}
\caption{Predicted depth maps generated by the proposed model for RGB images with added noise, blur, and occlusion under indoor (a) and outdoor (b) scenarios.}
\label{fig:image_processing}
\end{figure}

We applied noise, blur, and occlusion to the same RGB image and then generated the predicted depth maps using our proposed trained model. This process was performed on both indoor and outdoor images. Figure \ref{fig:image_processing} shows the results for each case, demonstrating the model's performance under varying conditions.

\section{Discussion}

The proposed model, which uses the Inception-ResNet-v2 architecture in an encoder-decoder framework, makes considerable improvements in monocular depth estimation. The combination of multi-scale feature extraction with a composite loss function has shown useful in addressing a wide range of depth estimation difficulties, attaining great accuracy and efficiency on benchmark datasets including KITTI and NYU Depth V2. These findings support the model's ability to generalize between indoor and outdoor environments, exceeding several cutting-edge architectures in a variety of criteria. However, some limits and practical constraints for real-world applications persist.

Our approach achieves lower accuracy than advanced depth prediction techniques such as Vision Transformer (ViT). Our model obtains an accuracy of 89.9\%, whereas the smallest variation of ViT (DINOv2-S) achieves 92.3\% for $\delta$ < 1.25. However, it outperforms ViT in terms of efficiency. Table \ref{table:kitti_table_inference} shows that our model achieves an inference time of 0.019 seconds for an input size of 240 × 320, while ViT-S requires 0.781 seconds on the same 2x NVIDIA T4 GPUs. Despite ViT-S's 1.02x higher accuracy, our model has a 41x faster inference time. Though increasing the input resolution improves output accuracy, it also increases inference time for Vision Transformers. In real-world applications, efficiency might sometimes take precedence over reaching the utmost precision, especially when the accuracy is already at an acceptable standard.

While the inference time of 0.019 seconds is enough for many applications, the model's relatively large parameter count may provide difficulties for deployment in resource-constrained situations, such as mobile devices or embedded systems. These issues can be addressed via optimization approaches such as model reduction, quantization, and efficient parallel computation. Furthermore, real-world settings frequently include differences in lighting, resolution, and noise levels, necessitating robust preprocessing pipelines and careful evaluation of a variety of datasets to assure consistent performance.

Despite these challenges, the proposed model's scalability and modularity make it adaptable to future hardware and algorithm developments. These characteristics make it ideal for applications such as robotics, augmented reality, and 3D reconstruction. Future research could concentrate on reducing computing complexity and increasing resilience to challenging environments, ensuring practicality and efficacy in deployment.

\section{Conclusion}
This research presents a noble method for depth map production utilizing monocular depth estimation through an encoder-decoder architecture founded on the Inception-ResNet-v2 model. Our methodology employs multi-scale feature extraction and a composite loss function that integrates depth loss, gradient edge loss, and SSIM loss, resulting in a notable improvement in depth prediction accuracy, evidenced by an Absolute Relative Error (ARE) of 0.064, a Root Mean Square Error (RMSE) of 0.228, and a Log10 error of 0.032. These values demonstrate the model’s precision in predicting accurate depth values for indoor scenarios. Besides, our model achieved 89.3\% accuracy for $\delta$ $<1.25$, outperforming other approaches in complex scenarios with changing object sizes and distances. Moreover, our model achieves significantly lower inference time compared to state-of-the-art Vision Transformer models (DINOv2 \cite{oquab2023dinov2}) on the KITTI \cite{dataset_KITTI} dataset, while maintaining an acceptable level of accuracy. These results demonstrate the potential of our model for real-time applications, where computational efficiency is crucial and maintaining good accuracy is essential.

Integration with other sensors and systems, such as GPS and LiDAR, can improve overall performance and reliability when deployed in real-time systems such as autonomous cars or drones. The model's energy consumption should also be tuned for battery-powered devices to ensure operational feasibility during long-duration tasks. Furthermore, compliance with ethical and regulatory standards, especially in applications such as surveillance or healthcare, is critical for public acceptance and legal use.

One drawback of our architecture is its higher computational cost (110 GFLOPs), which makes the model more resource-intensive and slower during both training and inference. In real-time applications, such as autonomous driving or robotic navigation, this could cause latency, limiting the model's applicability in resource-constrained environments like mobile devices. However, it still significantly outperforms state-of-the-art transformer-based methods in terms of efficiency.

Despite these challenges, the model's improvements in error metrics and high accuracy make it a promising solution for real-world applications. Offering more reliable depth predictions in complex environments, it performs well in generating accurate depth maps even in intricate and demanding situations. Compared to Vision Transformers, our model demonstrates superior efficiency, making it particularly well-suited for critical tasks, such as autonomous systems and other real-time applications, where reliability and performance are crucial.

\section*{Declaration of Funding}
This paper was not funded. 


\section*{Author Contributions}
\textbf{Dabbrata Das:} Conceptualization, Methodology, Visualization, Writing – original draft. \textbf{Argho Deb Das:}  Conceptualization, Formal analysis, Writing – original draft \& review. \textbf{Farhan Sadaf:} Formal analysis, Investigation, Writing – review \& editing, Project administration.

\section*{Ethical Approval}
Not required.

\section*{Declaration of Competing Interest}
The authors declare no competing interests.

\section*{Acknowledgements}
This work is supported in part by the Khulna University of Engineering \& Technology (KUET).



\bibliographystyle{model1-num-names}


\end{document}